\documentclass{article} % For LaTeX2e
\usepackage{collas2025_conference,times}
\usepackage{easyReview}
\usepackage{xcolor}
\usepackage{graphicx}
\usepackage{subcaption}
\usepackage[utf8]{inputenc}
\usepackage{booktabs} % for \toprule, \midrule, \bottomrule
\usepackage{caption}  % for \caption in sub-tables (if needed)
\usepackage{subcaption} % for sub-tables, if you prefer

\usepackage{float}
\definecolor{customblue}{RGB}{0, 102, 204}   % Blue for Author One
\definecolor{customred}{RGB}{204, 0, 0}      % Red for Author Two
\definecolor{customgreen}{RGB}{0, 153, 76}   % Green for Author Three

\newcommand{\equalcontrib}{\textsuperscript{*}}

\usepackage{amsmath,amsfonts,bm}

\def\eqref#1{equation~\ref{#1}}
\def\1{\bm{1}}

\def\ra{{\textnormal{a}}}

\def\rx{{\textnormal{x}}}

\def\rva{{\mathbf{a}}}

\def\erva{{\textnormal{a}}}

\def\ervx{{\textnormal{x}}}

\def\rmA{{\mathbf{A}}}

\def\vmu{{\bm{\mu}}}
\def\vtheta{{\bm{\theta}}}
\def\va{{\bm{a}}}

\def\ve{{\bm{e}}}

\def\vx{{\bm{x}}}

\def\eva{{a}}

\def\mA{{\bm{A}}}

\def\mH{{\bm{H}}}
\def\mI{{\bm{I}}}
\def\mJ{{\bm{J}}}

\def\mX{{\bm{X}}}

\def\mSigma{{\bm{\Sigma}}}

\DeclareMathAlphabet{\mathsfit}{\encodingdefault}{\sfdefault}{m}{sl}
\SetMathAlphabet{\mathsfit}{bold}{\encodingdefault}{\sfdefault}{bx}{n}
\newcommand{\tens}[1]{\bm{\mathsfit{#1}}}
\def\tA{{\tens{A}}}

\def\tX{{\tens{X}}}

\def\gG{{\mathcal{G}}}

\def\sA{{\mathbb{A}}}
\def\sB{{\mathbb{B}}}

\def\sR{{\mathbb{R}}}
\def\sS{{\mathbb{S}}}

\def\emA{{A}}

\newcommand{\etens}[1]{\mathsfit{#1}}

\def\etA{{\etens{A}}}

\newcommand{\E}{\mathbb{E}}

\newcommand{\R}{\mathbb{R}}

\newcommand{\KL}{D_{\mathrm{KL}}}
\newcommand{\Var}{\mathrm{Var}}

\newcommand{\Cov}{\mathrm{Cov}}
\newcommand{\normltwo}{L^2}
\newcommand{\normlp}{L^p}

\newcommand{\parents}{Pa} % See usage in notation.tex. Chosen to match Daphne's book.

\usepackage{hyperref}
\hypersetup{
    colorlinks=true,
    linkcolor=red,
    filecolor=magenta,
    urlcolor=blue,
    citecolor=purple,
    pdftitle={Overleaf Example},
    pdfpagemode=FullScreen,
    }

\title{Preserving Plasticity in Continual Learning with Adaptive Linearity Injection}
\author{Seyed Roozbeh Razavi Rohani\equalcontrib \\
Department of Computing Science\\
Simon Fraser University\\
Burnaby, BC, Canada \\
\texttt{srr8@sfu.ca} \\
\And % Use And to have authors side by side
Khashayar Khajavi\equalcontrib \\
Department of Computing Science \\
Simon Fraser University \\
Burnaby, BC, Canada \\
\texttt{kka151@sfu.ca} \\
\And % Use AND to have authors block one under the other
Wesley Chung \\
Mila, McGill University \\
Montreal, QC, Canada \\
\texttt{chungwes@mila.quebec} \\
\AND
Mo Chen \\
Department of Computing Science \\
Simon Fraser University \\
Burnaby, BC, Canada \\
\texttt{mochen@cs.sfu.ca} \\
\And
Sharan Vaswani \\
Department of Computing Science \\
Simon Fraser University \\
Burnaby, BC, Canada \\
\texttt{vaswani.sharan@gmail.com}
}

\preprintcopy % Uncomment for the preprint version, but NOT for submission.

\begin{document}

\maketitle
\begingroup
  \renewcommand\thefootnote{\textasteriskcentered}%
  \footnotetext{These authors contributed equally to this work.}
\endgroup
\begin{abstract}
% Deep neural network architectures have been typically designed under the assumption of stationary data, yet real-world continual learning scenarios expose them to non-stationary environments that lead to a loss of plasticity—the gradual reduction in a model’s capacity to acquire new knowledge.
Loss of plasticity in deep neural networks is the gradual reduction in a model’s capacity to incrementally learn and has been identified as a key obstacle to learning in non-stationary problem settings. Recent work has shown that deep linear networks tend to be resilient towards loss of plasticity. Motivated by this observation, we propose \textbf{Ada}ptive \textbf{Lin}earization (\texttt{AdaLin}), a general approach that dynamically adapts each neuron's activation function to mitigate plasticity loss.  Unlike prior methods that rely on regularization or periodic resets, \texttt{AdaLin} equips every neuron with a learnable parameter and a gating mechanism that injects linearity into the activation function based on its gradient flow. This adaptive modulation ensures sufficient gradient signal and sustains continual learning without introducing additional hyperparameters or requiring explicit task boundaries. When used with conventional activation functions like ReLU, Tanh, and GeLU, we demonstrate that \texttt{AdaLin} can significantly improve the performance on standard benchmarks, including Random Label and Permuted MNIST, Random Label and Shuffled CIFAR 10, and Class-Split CIFAR 100. Furthermore, its efficacy is shown in more complex scenarios, such as class-incremental learning on CIFAR-100 with a ResNet-18 backbone, and in mitigating plasticity loss in off-policy reinforcement learning agents. We perform a systematic set of ablations that show that neuron-level adaptation is crucial for good performance, and analyze a number of metrics in the network that might be correlated to loss of plasticity.
\end{abstract}

\section{Introduction}

% Continual learning studies how an agent adapts to a sequence of tasks while encountering an ever-changing distribution of new data, and is widely used in different domains ~\citep{van2019three,wang2024comprehensive}. Unlike humans and other biological agents that excel at lifelong learning, deep learning models struggle due to plasticity loss, a gradual decline in their ability to learn and adapt.
Loss of plasticity in neural networks refers to the gradual decline in a model’s ability to learn and adapt to new data over time, a phenomenon that is particularly exacerbated in non-stationary learning settings such as continual learning and reinforcement learning~\citep{nikishin2023deep,lyle2023understanding,lyle2024disentangling,dohare2024loss}.
To address this challenge, recent approaches have mostly focused on regularizing the network during training~\citep{kumar2023maintaining,lewandowski2023curvature, chung2025parseval} or periodically reinitializing a subset of weights to restore learning capacity ~\citep{ash2020warm,dohare2024loss}. While these methods can alleviate plasticity loss in controlled settings, they often come with
limitations. Regularization can overly constrain the model, reducing its overall capacity and performance, while reinitialization strategies often depend on explicit task boundaries and can disrupt learned representations, making them less suitable for real-world applications where tasks evolve continuously.

\begin{figure}[H]
    \centering
    % First subfigure
    \begin{subfigure}[b]{0.48\linewidth}
        \centering
        \includegraphics[width=\linewidth]{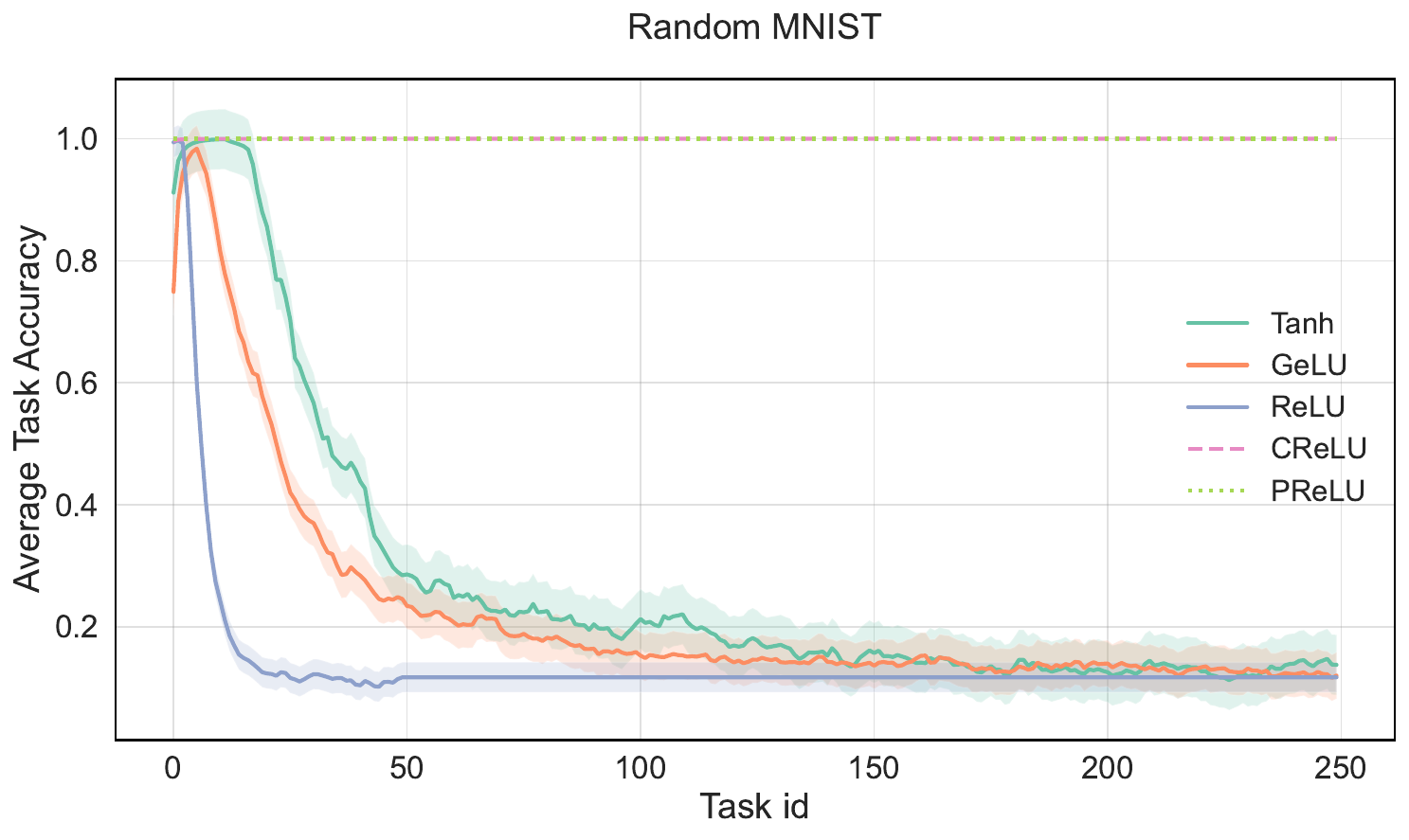}
        \caption{The impact of different activation functions on plasticity loss on the random-label MNIST task.
        Activation functions such as CReLU and PReLU help sustain the network’s learning capacity over multiple tasks, 
        while others experience significant degradation. This highlights the crucial role of activation function choice 
        in maintaining plasticity and long-term learning ability in neural networks.}
        \label{fig:toy_example}
    \end{subfigure}
    \hfill
    % Second subfigure
    \begin{subfigure}[b]{0.48\linewidth}
        \centering
        \includegraphics[width=\linewidth]{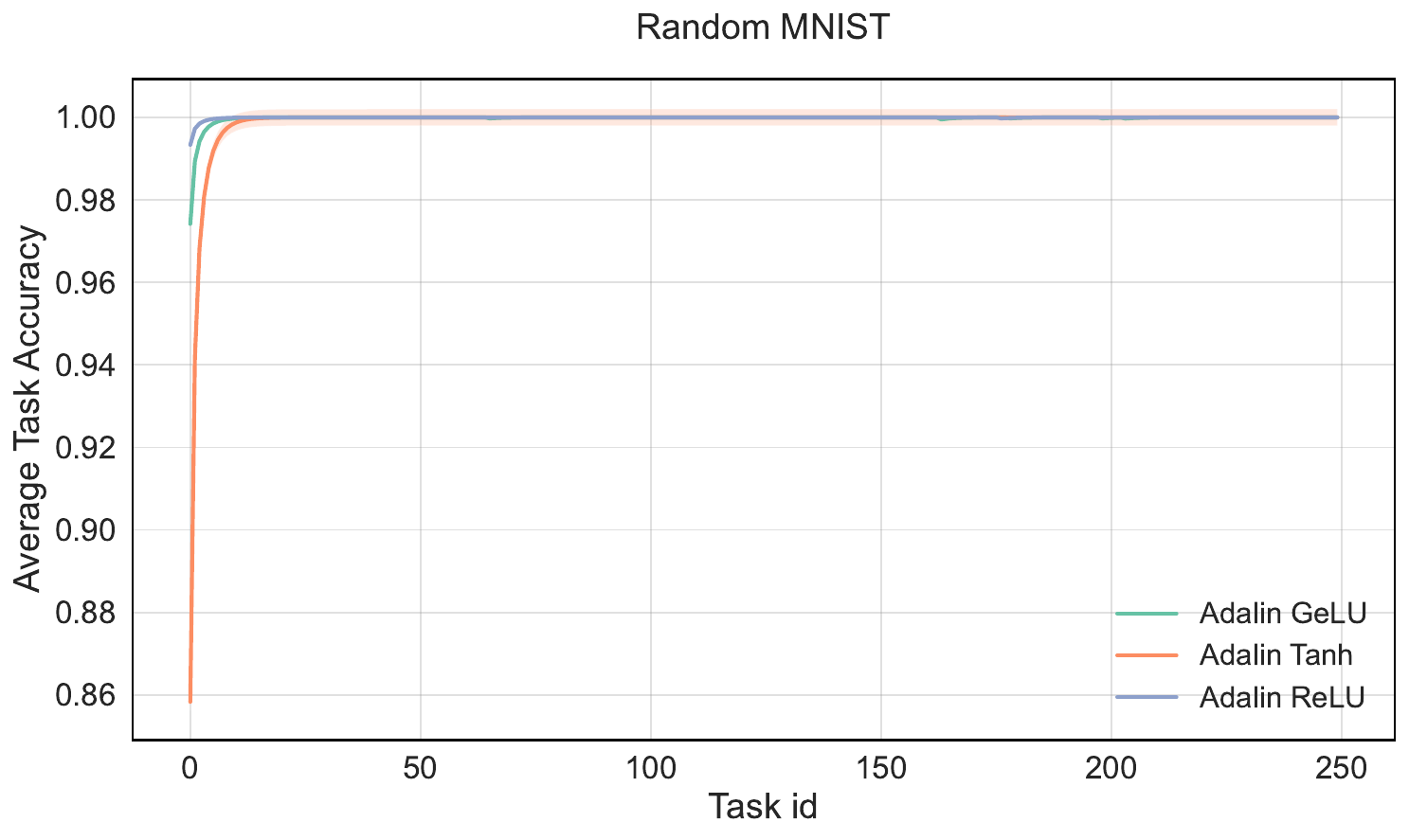}
        \caption{\texttt{AdaLin}, our proposed method, adaptively injects linearity into the network and and can be used in conjunction with any given activation function. The figure shows the continual learning performance corresponding to using \texttt{AdaLin} with common activation functions such as Tanh, ReLU, GeLU. Comparing to Figure \ref{fig:toy_example}, we observe that \texttt{AdaLin} can help sustain high task accuracy over successive tasks in the random-label MNIST scenario. 
        }
        \label{fig:toy_example_improved}
    \end{subfigure}
    \caption{(a) Baseline comparison of activation functions and their vulnerability to plasticity loss in the random-label MNIST task. 
    (b) Improved performance when combining \texttt{AdaLin} with these activations.}
    \label{fig:toy_example_sidebyside}
\end{figure}

Consequently, an alternative line of research has investigated how activation functions influence plasticity loss and highlighted their central role in sustaining a model’s learning ability over time \citep{abbas2023loss,lyle2024disentangling,lewandowski2024plastic}. A straightforward example of this phenomenon appears in Figure~\ref{fig:toy_example}, where the same MLP trained on the Random-label MNIST task\footnote{In this benchmark, 1200 random images are chosen from the MNIST dataset. For each task, every image is assigned a new random label. The experiment measures continual learning performance (in terms of average task accuracy) across a sequence of tasks, with each task representing one complete training cycle on the randomized labels.} exhibits drastically different outcomes depending on its activation function. Specifically, standard activations such as ReLU, GeLU~\citep{hendrycks2016gaussian}, and Tanh fail to sustain the network’s learning capacity, experiencing significant degradation over time. By contrast, alternatives like CReLU~\citep{abbas2023loss} and PReLU~\citep{he2015delving} better maintain plasticity across tasks. CReLU achieves this by concatenating the positive and negative responses, ensuring that a portion of each neuron's output remains unsaturated. PReLU, on the other hand, introduces a learnable slope for negative inputs. Recent findings also suggest that deep linear networks, where the activation function is simply the identity, are more resilient to loss of plasticity, but lack the expressive power of nonlinear functions .\citep{lewandowski2023curvature,lewandowski2024plastic,dohare2024loss}. This observation has inspired methods such as deep Fourier features~\citep{lewandowski2024plastic}, which embed a deep linear model into the network by outputting both sine and cosine components \footnote{A detailed description of the deep Fourier method can be found in Appendix \ref{sec:baselines}.}. 

%This approach combine the advantages of linearity with the expressive power of nonlinearity, mitigating plasticity loss and underscoring the importance of activation design for continual learning.

% Such an activation function one of which is always linear with respect to the input, aiming to embed a deep linear model within the network.

\begin{figure}[htb]
    \centering
    \includegraphics[width=0.6\linewidth]{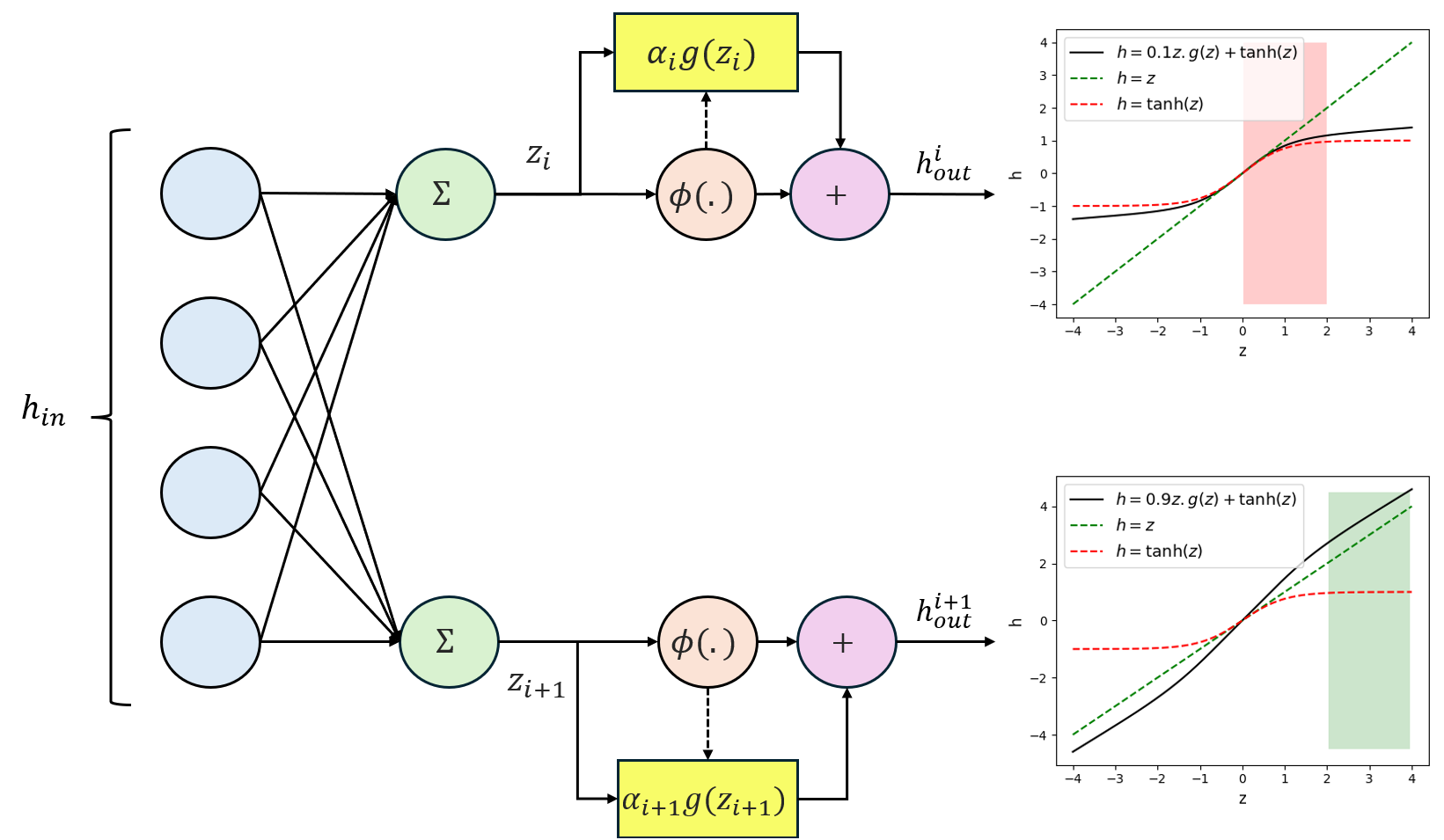}
    \caption{Illustration of \texttt{AdaLin}’s adaptive activation mechanism. (Left) A neural network structure where
    a parameter $\alpha$ is learned per each neuron that modulates its activation between linear and non-linear components.
    Here, ${h}^i_{\text{out}}$ and $z_i$ represent the post-activation and pre-activation values of neuron $i$, respectively.
    $\phi(\cdot)$ is a base activation function, and $g(\cdot)$ is a gating function computed from the gradient of $\phi(\cdot)$
    (i.e., it automatically adjusts based on how close $\phi(\cdot)$ is to saturation). (Right)
    Plots demonstrating the behavior of the learned activation function, where \texttt{AdaLin} smoothly interpolates between
    a standard nonlinearity (e.g., $\tanh(x)$, red) and a linear function (green). This injection of linearity is crucial in
    saturated regions (where the derivative of $\phi(\cdot)$ is small), helping sustain gradient flow and preserve plasticity.
    In the top figure, $z_i$ falls in the non-saturated region of $\tanh(x)$ (red), so no linearity is added. In the bottom
    figure, $z_{i+1}$ lies in a saturated region (green), prompting the gating function $g(\cdot)$ to enable linear injection
    based on $\alpha_{i+1}$.}
    \label{fig:overall}
\end{figure}

    \vspace{-4mm}
    Deep Fourier features effectively embed linear components by mapping each preactivation to two outputs, ensuring that, at any point, one of the two responses of a neuron is approximately linear. However, this fixed ratio of linear to non-linear outputs may not be optimal across different tasks. Motivated by the need to adaptively adjust this balance, in this work, we propose \textbf{Ada}ptive \textbf{Lin}earization (\texttt{AdaLin}), a simple yet effective method for dynamically balancing linear and non-linear behavior in neural network activations. With \texttt{AdaLin}, each neuron learns a \textit{linearity injection} parameter that determines the extent of the linear component added to its base activation. This adaptive mechanism allows neurons to modulate the proportion of linearity in their response based on how close their activation is to the saturation region. This balance between the stability of linear models and the expressive power of non-linear activations fosters continual adaptation without losing plasticity across tasks. Figure \ref{fig:overall} depicts an illustration of how \texttt{AdaLin} functions.

    Furthermore, biological studies highlight the importance of dynamic adaptability in neural responses. Neuroscience research shows that, unlike fixed activation functions in conventional deep learning models, biological neurons dynamically adjust their response profiles in reaction to contextual factors and neuromodulatory influences~\citep{Ferguson2020,Cheadle2014,McGinley2015}.  A more comprehensive review and related neuroscientific findings on this topic can be found in Appendix~\ref{sec:neuro_inspired}. While numerous studies ~\citep{apicella2021survey, he2015delving, trottier2017parametric, qiu2018frelu} have explored activation functions that can be trained during the learning process, the absence of these dynamic methods in continual learning may contribute to challenges like loss of plasticity. 

We evaluate \texttt{AdaLin} on multiple plasticity benchmarks~\citep{kumar2023maintaining,galashov2025non} and show that it consistently matches or outperforms recent methods, all without introducing extra hyperparameters or relying on explicit task boundaries. Moreover, compared to other activation functions like CReLU and deep Fourier features, this approach shows more robust performance across benchmark tasks and does not need to output two activations per neuron. Our method demonstrates that incorporating a single learnable parameter per neuron can substantially help preserve plasticity in continual learning. By analyzing the learned injection parameters, we observe that neurons nearing saturation tend to inject a small negative linear component, whereas those that remain unsaturated predominantly rely on their non-linear behavior. Notably, as a special case, \texttt{AdaLin} with ReLU recovers the well-known PReLU formulation—traditionally designed for single-task learning—highlighting its novel application in continual learning settings. Extensive ablation studies across ReLU, Tanh, and GELU confirm that \texttt{AdaLin} not only mitigates plasticity degradation but also improves training stability and generalization across a wide range of tasks.

%neurons that approach saturation inject a small linear component, whereas those that remain unsaturated rely more on their nonlinear behavior.
% An example of this attribute can be seen in Figure \ref{fig:toy_example_improved}.

% Our results show that in most tasks, only a small fraction of neurons adopt a linear response
%\input{sections/2-related_work}
% \input{sections/3-problem_setting}
% \input{sections/4-toy_example}
\section{Adaptive Linearization}\label{sec:adalin}
\iffalse
\subsection{Notation}

We formalize our problem setting in a non-stationary learning environment where the data distribution at time \( t \) is given by \( p_t(x, y) \) with \( x \in \mathbb{R}^L \) and \( y \in \mathbb{R}^K \). For parameters \( \theta \in \mathbb{R}^D \), the loss function at time \( t \) is defined as
\[
\mathcal{L}_t(\theta)=\mathbb{E}_{(x_t, y_t) \sim p_t} \mathcal{L}_t\left(\theta, x_t, y_t\right).
\]

Our objective is to find a sequence of parameters \(\Theta=\left(\theta_1, \ldots, \theta_T\right)\) that minimizes the dynamic regret
\[
R_T\left(\Theta, \Theta^{\star}\right)=\frac{1}{T} \sum_{t=1}^T\left(\mathcal{L}_t\left(\theta_t\right)-\mathcal{L}_t\left(\theta_t^{\star}\right)\right),
\]
where the reference sequence \(\Theta^{\star}=\left(\theta_1^{\star}, \ldots, \theta_T^{\star}\right)\) is defined by 
\[
\theta_t^{\star}=\arg \min _\theta \mathcal{L}_t(\theta).
\]
This formulation provides the framework for our investigation into strategies that maintain learning plasticity over extended sequences of tasks.
\fi

% \subsection{Dynamic Adaptation of Linearity (DALin)}

To introduce the details of \texttt{AdaLin}, we first consider a multi-layer perception network using a fixed, Lipschitz activation function $\phi: \sR \rightarrow \sR$ in all hidden layers $\ell \in \{1, ..., L\}$, where $L$ is the number of hidden layers. This includes many common activation functions such as those in the ReLU family (ReLU, GELU, Mish~\citep{misra2019mish}) or sigmoidal functions (sigmoid, Tanh). We later describe how \texttt{AdaLin} can be generalized to other network architectures than MLPs. 

We use the variable $i$ to index over all hidden units in the network. So, $i \in \{1, ..., N_{neurons}\}$ where $N_{neurons}$ is the total number of neurons in the network (including across hidden layers). 
We let $h_{in}^i$ denote the vector of inputs to neuron $i$, $h_{out}$ denote the neuron's scalar output and $w_i$ denote the neuron's learned weights. 
During a forward pass of the network, every neuron computes
\begin{equation} h_{out}^i = \phi(w_i \cdot h_{in}^i) \end{equation}

\texttt{AdaLin} is an approach to augment the base activation function $\phi$ by adding a linear term in the input and a learnable parameter modulating between $\phi$ and the linear part. 
Concretely, for neuron $i$, given an activation function $\phi: \sR \rightarrow \sR$, we define a new activation function $f:\sR \rightarrow \sR$ as follows:
\begin{equation}\label{eq:DALIN}
f_i(x) = \phi(x) + \alpha_i\, x \; [ g(x)]_{sg}, \quad g(x) = \cos\left(\frac{\pi}{2} \cdot \frac{|\phi'(x)|}{L}\right).
\end{equation}
where $\alpha_i$ is a learnable \textit{linearity injection} parameter, $\phi$ is a nonlinear, Lipschitz function (e.g., Tanh or ReLU) and  $[f]_{sg}$ indicates a stop-gradient on $f$ so backpropagation considers $f$ as a constant. 
Also, $L$ represents the Lipschitz constant of $\phi(x)$, and $\phi'(x)$ denotes the first derivative of $\phi$ evaluated at $x$.

\begin{figure}[htb]
    \centering
    \includegraphics[width=0.5\linewidth]{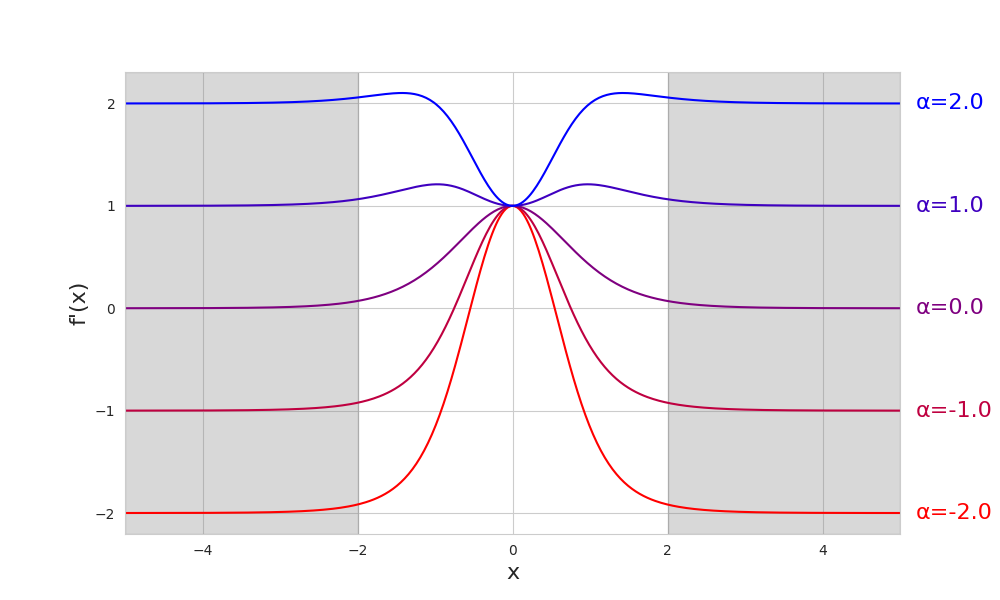}
    \caption{Derivative of \texttt{AdaLin} with $\tanh$ as the base activation function under varying $\alpha$. 
    For $\alpha=0$, this gives the original $\tanh$ function and we see that when x lies in the saturated region of $\tanh$
(shaded in gray), the derivative is zero, limiting the gradient flow through the corresponding neuron.
As $\alpha$ takes a non-zero value, a linear term is injected, ensuring a nonzero derivative for all possible input values.}

    \label{fig:f_prime}
\end{figure}

This modified activation function has a derivative
\[f'_i(x) = \phi'(x) + \alpha_i g(x) \]
with respect to its input.
From this expression, we can see that $g(x)$ acts as a gating mechanism, controlling the contribution of the linear term to the output $f_i(x)$.
The function $g(x)$ is bounded between $0$ and $1$ since $\sup_{x \in \sR} |\phi'(x)| \le L$ wherever the derivative exists.  
The gating is modulated by the magnitude of the derivative of the base activation $\phi'(x)$. 
When the magnitude is small, then we can expect difficulties in optimization due to the small (or zero) gradients~\citep{abbas2023loss,dohare2024loss, sokar2023dormant}.
In this case, $g(x)$ is near 1 and the linear term contributes a larger derivative to obtain a learning signal. 
Conversely, when $\phi'(x)$ is large, there is no issue and thus $g(x)$ is near zero. This is summarized in Figure \ref{fig:f_prime}. All in all, $g(x)$ aims to provide a measure on the \textit{Saturation Rate} of $\phi(x)$ (i.e., how saturated $\phi(.)$ is in the proximity of the input $x$).
Furthermore, an illustration of \texttt{AdaLin}  is provided in Figure \ref{fig:overall}.

% To analyze the behavior of Equation \ref{eq:DALIN}, we consider what happens as the norm of $\phi'(x)$ is changing.

% WES: Add a plot g(x) with f(x) side-by-side

% WES: Add discussion of stop gradient

The design of the gating function $g(x)$ can be modified as long as its core components are respected: $g(x)$ takes value zero wherever the derivative $\phi'(x)$ is maximal and takes value near one when $\phi'(x)$ is near zero. 
We propose the formulation in equation \ref{eq:DALIN} as a simple example fitting these requirements. Moreover, we compare the performance of different gating functions that satisfy these requirements in Appendix \ref{appendix:lin}.
Note that the stop-gradient operation is utilized to ensure that the derivative of $g(x)$ does not adversely influence the derivative of $f_i$.

\texttt{AdaLin} can be easily applied to other network architectures beyond MLPs. Since the approach modifies the activation on a per-neuron basis, it can also be applied to each neuron of other types of neural network layers. For example, for convolutional layers that utilize weight-sharing between different neurons, we can apply weight sharing in a similar fashion for the learnable parameters $\alpha_i$ of \texttt{AdaLin}. A more detailed description appears in Appendix \ref{sec:appendix_cnn}. Moreover, as shown in Appendix \ref{appendix:overhead}, the proposed method adds negligible overhead to the learnable parameters in the network.

The proposed algorithm builds upon the idea of introducing adaptive linearity in activation functions, as demonstrated in Parametric ReLU (PReLU)~\citep{he2015delving}, and generalizes this concept to work with any Lipschitz activation function.
While PReLU adjusts ReLU by adding a learnable linear component for $x < 0$, the presented approach extends this principle by dynamically injecting a tunable linear component into any activation function wherever the gradient is near zero.
Specifically, as shown in Appendix \ref{sec:appendix_prelu}, when $\phi(.)$ is ReLU, the method recovers the PReLU formulation. 
Since PReLU was originally proposed for the single-task setting, it has not been explored in continual learning settings. We find that it can be effective for maintaining plasticity, a novel benefit of this activation function.

%Moreover, instead of being limited to a predefined modification of a specific activation function, our method provides a systematic framework that adapts non-linearities based on their saturation behavior.
% By integrating this adaptive mechanism, \texttt{AdaLin} enhances the flexibility of activation functions, allowing them to maintain useful transformations while mitigating the loss of plasticity caused by saturation. This adaptation ensures that neurons can better retain their learning capacity over time, making activation functions more effective in continual learning settings.

We analyze the impact of \texttt{AdaLin} in Section~\ref{exp:plastic-benchmarks}, where we demonstrate its effectiveness across different activation functions and learning scenarios.
Through our experiments and ablation studies, we evaluate its use with various activation functions, including Tanh, Gaussian Error Linear Units (GELU), and ReLU. 
Our results empirically demonstrate that combining a gating mechanism with adaptive scaling can significantly improve plasticity for every activation function $\phi$ investigated. 
Additionally, in Section~\ref{exp:neuron-behavior}, we analyze the learned $\alpha$ values and the gating function $g(x)$ across neurons to gain insight into their evolution over different tasks and their role in balancing plasticity and stability in continual learning. 
Furthermore, our experiments in Appendix \ref{apendix:gx} demonstrate that incorporating the gating function $g(x)$ is critical for optimal performance, as it dynamically modulates the linearity injection and ensures that the residual connection is activated only when the neuron is saturated.

\vspace{-2mm}
\section{Experiments}\label{sec:experiments}
\vspace{-2mm}

In this section, we use continual supervised learning settings to empirically evaluate the proposed method. Following \cite{dohare2024loss, kumar2023maintaining, galashov2025non} we focus on a scenario where a learning agent sequentially encounters several distinct tasks. The agent processes the data for each task incrementally, receiving and learning from batches of data over multiple timesteps in every task.
% Through this comparative analysis, we highlight the effectiveness of this approach in maintaining adaptability while learning new tasks.

\subsection{Plasticity Benchmarks}
\vspace{-2mm}

% In our investigation of plasticity loss, we consider five benchmark problems. These benchmarks explore a variety of task change. 
% In the first set of tasks, Random Label MNIST and Random Label CIFAR-10, we have 

In our investigation of plasticity loss in neural networks, we evaluate our approach to a variety of tasks, each introducing non-stationarity in a different way. Some tasks randomly reassign labels, forcing the network to memorize arbitrary mappings. Others transform the input data, for example, by permuting image pixels. Finally, tasks like class-split benchmarks gradually introduce new classes, shifting both the input distribution and the label space. Each experiment thus represents a distinct mechanism for non-stationarity—ranging from label reassignments to input permutations and class increments—and offers a broad perspective on how plasticity loss can manifest under different distributional changes. A more detailed description of these benchmarks is provided in Appendix~\ref{appendix:benchmarks}.

\begin{figure}[!t] % [h] means 'here' (preferred position)
    \centering \includegraphics[width=0.8\textwidth]{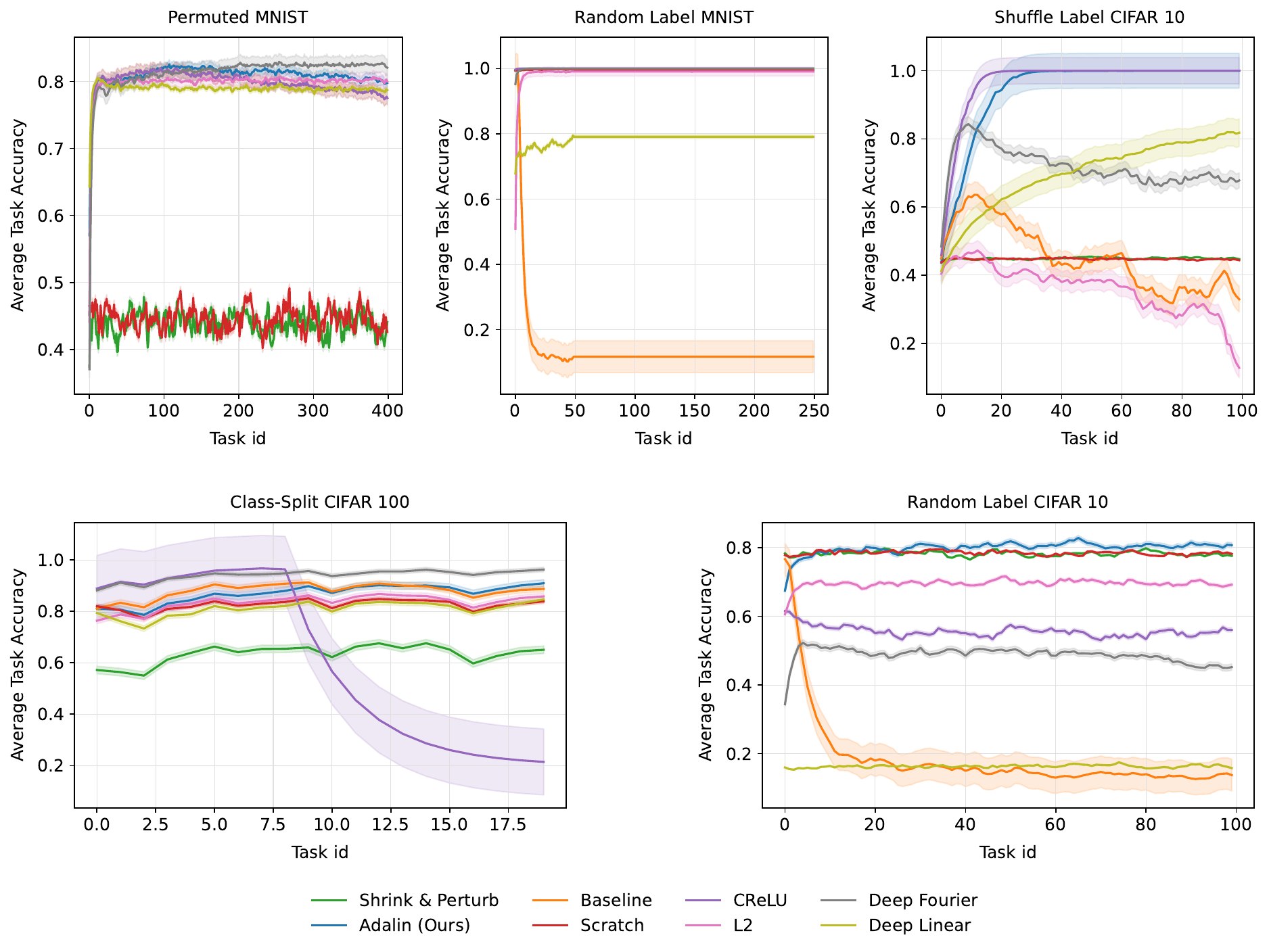}
    \caption{Average Task Accuracy across five plasticity benchmarks (Permuted MNIST, Random Label MNIST,
Shuffle Label CIFAR-10, Class-Split CIFAR-100, and Random Label CIFAR-10) for various methods, all using
ReLU as the base activation except the \texttt{Deep Linear} model. Overall, \texttt{AdaLin} consistently maintains plasticity
and often outperforms other approaches, particularly in tasks where transferring knowledge can improve learning
(e.g., Permuted MNIST, Shuffled Label CIFAR-10, and Class-Split CIFAR-100). Methods like \texttt{Scratch} and \texttt{Shrink \& Perturb}
struggle due to frequent resetting, while baseline ReLU models degrade severely in random-label tasks. Although
\texttt{Deep Linear} does not exhibit plasticity loss, its limited expressivity leads to lower overall performance.}

    \label{fig:result_train}
\end{figure}

\vspace{-3mm}
\subsection{Evaluation and Training Setup}
\vspace{-2mm}

To evaluate the model's performance and its ability to maintain plasticity, we formalize the continual learning scenario: a learning agent is presented with a sequence of \(K\) distinct tasks, where each task \(T_i\) is associated with a dataset \(\mathcal{D}_{T_i}\) of labeled examples (i.e., images). The agent processes each dataset incrementally in batches over \(M\) timesteps, allowing us to track its performance over time. Moreover, following the approach in previous works \citep{kumar2023maintaining}, we compute and track the average online train accuracy for each task. Specifically, for a given task $\mathrm{T}_i$, 

\vspace{-4mm}
\[
\text{Avg. Online Train Accuracy} (\mathrm{T}_i) = \frac{1}{M} \sum_{j=t_i}^{t_i+M-1} a_j
\]
\vspace{-4mm}

where $t_i$ represents the starting time step of task $\mathrm{T}_i$, and $a_j$ denotes the accuracy obtained on the $j$-th batch of data samples. Assuming all tasks have comparable difficulty, a decline in the average online train accuracy indicates a loss of plasticity. 

% This metric, referred to as average online train accuracy, reflects how efficiently the agent learns a task. 

% If this value declines over time, it indicates plasticity degradation, 

In addition to evaluating performance on training data, we also report test accuracy for each task to capture both aspects of plasticity loss: the network’s diminishing ability to fit new data and its declining capacity to generalize to unseen examples ~\citep{dohare2024loss,lee2024slow} which plays a critical role in tasks like Permuted MNIST, Shuffled Label CIFAR-10, and Class-Split CIFAR-100.

We used the SGD optimizer for all experiments and performed a hyperparameters sweep over three seeds, exploring various learning rates and batch sizes for each method–benchmark pair. The configuration that maximized the average online accuracy was then selected. After this, each experiment was run on three additional seeds, and we report the mean outcome, with shaded regions in our plots representing one standard deviation.  For initializing the $\alpha$ values, we sampled uniformly between 0 and 1. All model architectures, hyperparameters, and other training details (including system setup) are described in Section~\ref{sec:train_setup}.

\vspace{-1mm}
\subsection{Baselines}\label{sec
:baselines}
\vspace{-2mm}

Since \texttt{AdaLin} aims to mitigate plasticity loss by adapting the activation functions, we primarily compare our method against existing techniques that also address plasticity loss through activation functions, while also incorporating a few baselines from the regularization and resetting families. For activation-based methods, we considered   Concatenated ReLU (\texttt{CReLU}) ~\citep{abbas2023loss} and \texttt{Deep Fourier} features ~\citep{lewandowski2024plastic}. It is important to note that both \texttt{CReLU} and \texttt{Deep Fourier} features output two values per neuron, effectively doubling the number of parameters in each layer compared to traditional single-output activations. We also consider a \texttt{Deep Linear} network with linear activations, which we expect to perform relatively poorly due to its lack of nonlinearity and limited expressivity, but not suffer from loss of plasticity.

% Additionally, we examined standard activation functions such as ReLU, Tanh, and GeLU in conventional neural networks which we called them Vanilla ReLU, Vanilla Tanh, and Vanilla GeLU respectively.

% Layer-Wise Parametric ReLU (LPReLU) assigns a learnable parameter to each layer.

In addition to activation-based methods, we compare our approach with \texttt{Shrink \& Perturb} ~\citep{ash2020warm}, a representative resetting-based method and using \texttt{$\ell_2$} regularization (\texttt{L2}), a representative regularization-based method. We also include two trivial baselines: a fully continual model (\texttt{Baseline}) and a model trained from scratch (\texttt{Scratch}) for every task, both with ReLU as their activation functions. We defer the detailed descriptions of each baseline to Appendix \ref{sec:baselines}.
% which resets all weights at the end of each task.

% Comparing our method with Vanilla ReLU and Deep Linear networks, which represent two extremes of our method, provides insight into how interpolating between these two approaches can affect performance. Additionally, we examine how our method performs against Deep Fourier, which similarly leverages linearity as a strategy to maintain plasticity. 

\begin{figure}[!t] % [h] means 'here' (preferred position)
    \centering
    \includegraphics[width=0.8\textwidth]{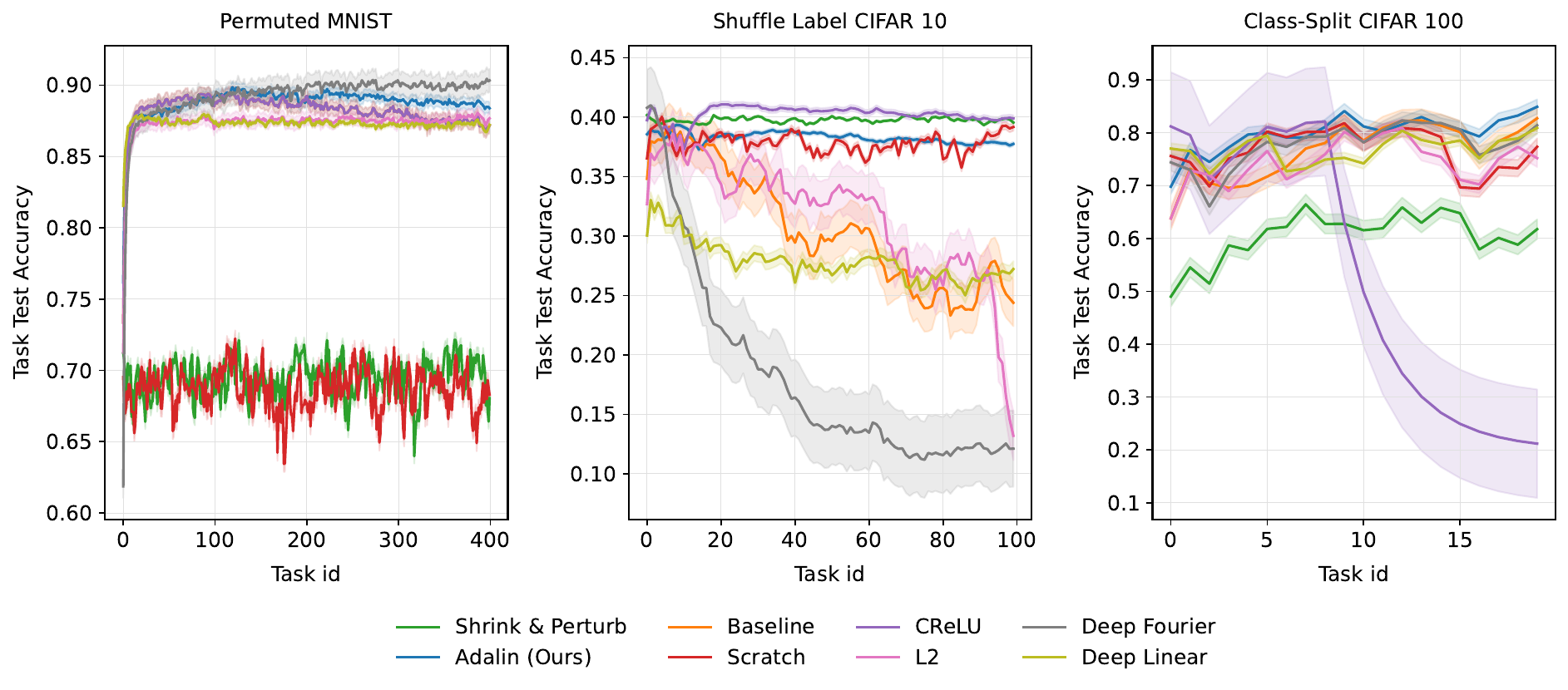}
    \caption{Test accuracy for tasks where generalization is particularly relevant (Permuted MNIST, Shuffle CIFAR-10,
and Class-Split CIFAR-100). \texttt{AdaLin} achieves the highest test accuracy in Class-Split CIFAR-100 and performs
strongly on Shuffle CIFAR-10 and Permuted MNIST. \texttt{Deep Fourier} struggles with generalization over time in Shuffle
CIFAR-10, while \texttt{Deep Linear}, despite avoiding plasticity loss in training, fails to generalize effectively in this setting.}
    \label{fig:result_test}
\end{figure}

\vspace{-1mm}
\subsection{Comparative Evaluation}
\label{exp:plastic-benchmarks}
\vspace{-2mm}
%seed + arch + hyper parameters
As shown in Figure \ref{fig:result_train}, \texttt{AdaLin} consistently maintains its plasticity across all benchmarks with the base activation function being ReLU for all the methods (note that when \texttt{AdaLin} is paired with ReLU, it recovers the well-known PReLU formulation.). In tasks where previously acquired knowledge could be transferred to improve learning, such as Permuted MNIST, Shuffled Label CIFAR-10, and Class-Split CIFAR-100, \texttt{AdaLin} can successfully leverage this advantage and even performed better than \texttt{Scratch}. In contrast, reset-based methods such as \texttt{Scratch} and \texttt{Shrink \& Perturb} performed poorly, resulting in lower average task accuracy. Additionally, all baseline methods experienced a loss of plasticity in at least one benchmark. Specifically, \texttt{CReLU} exhibits significant plasticity loss after approximately 10 tasks in Class-Split CIFAR-100, leading to nearly random performance. Importantly, the \texttt{Baseline} suffers from severe plasticity loss in Random Label MNIST and Random Label CIFAR-10. Moreover, in Shuffled Label CIFAR-10, the \texttt{Deep Fourier} and \texttt{$\ell_2$} regularization methods struggle to preserve plasticity. Although the \texttt{Deep Linear} model did not suffer from plasticity loss in any benchmark, its overall performance remained significantly lower than other methods due to the limited capacity of the linear network.

Beyond Average Task Accuracy, we also evaluated test accuracy to assess the test accuracy of different methods in tasks where generalization is meaningful (i.e. Permuted MNIST, Shuffle CIFAR-10, and Class-Split CIFAR-100) As shown in Figure \ref{fig:result_test}, our method achieved the highest test accuracy for CIFAR-100 and demonstrated strong performance on Shuffled Label CIFAR-10 and Permuted MNIST. However, the \texttt{Deep Fourier} method faces challenges in generalization, with its test accuracy dropping significantly over time in Shuffled Label CIFAR-10. Furthermore, while the \texttt{Deep Linear} network did not exhibit plasticity loss in training accuracy for this benchmark, its test accuracy declined, meaning that it cannot learn generalizable representations in a continual learning setting.

\subsection{Class Incremental Learning}
To further demonstrate the generality and real-world applicability of \texttt{AdaLin}, we evaluate its performance on a more complex, scaled architecture on the Class Incremental CIFAR-100 task with a ResNet-18 network, following the setup in ~\citet{lewandowski2024plastic, dohare2024loss}. Class-incremental learning involves introducing new classes over multiple tasks, testing the model’s ability to learn
from novel categories without forgetting previously learned ones. Following~\citet{lewandowski2024plastic}, we partition the
CIFAR-100 dataset into a series of tasks, each contributing five new classes to the training dataset. At the end of each task, the network is evaluated on the complete test set. As training progresses and new tasks are introduced, the network's test performance steadily improves because it is exposed to an increasing portion of the training data. This
approach offers a more realistic challenge for continual learning, as it requires balancing stability and adaptability on incremental  classes.

\begin{figure}[htb] % [h] means 'here' (preferred position)
    \centering
    \includegraphics[width=0.6\linewidth]{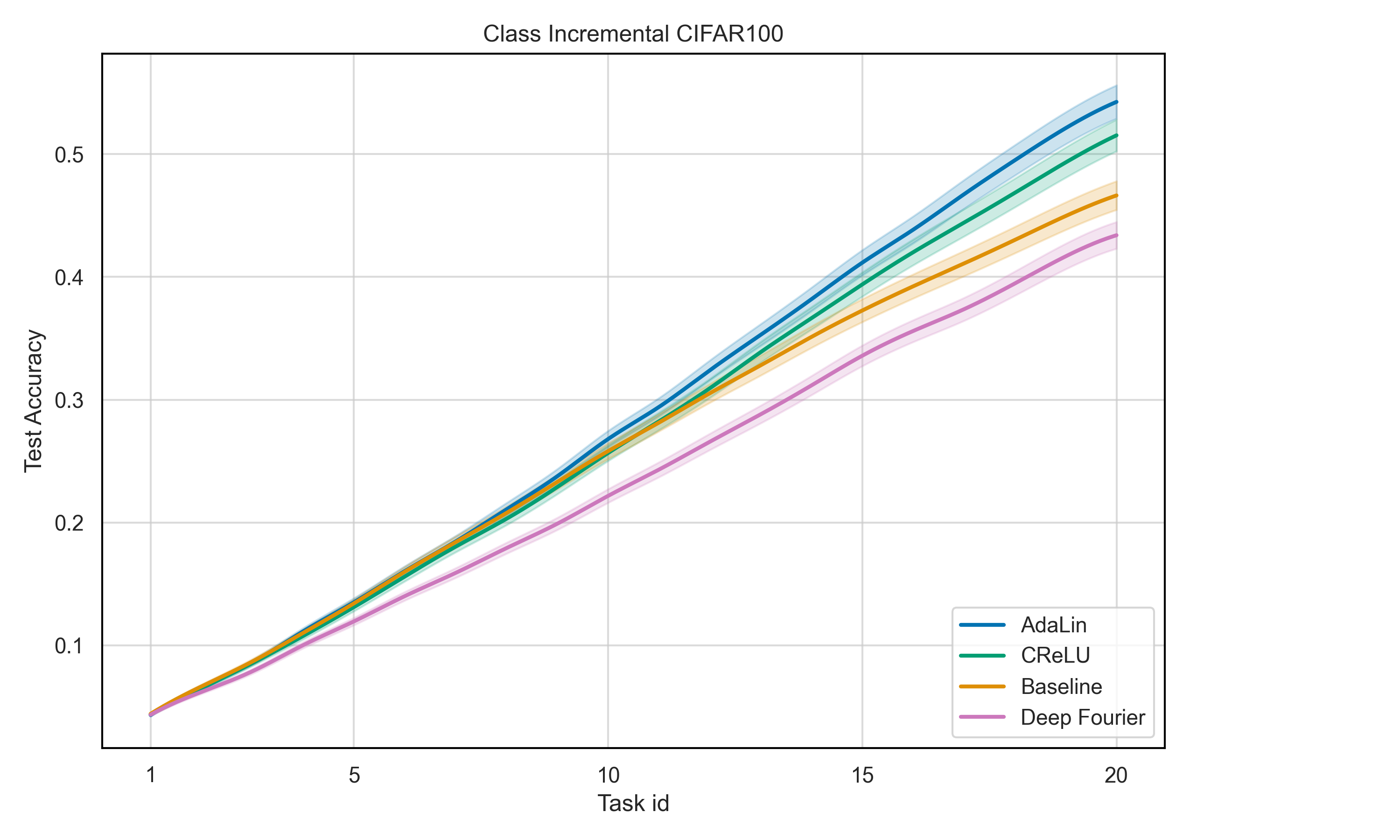}
    \caption{Class-incremental CIFAR-100 results using a ResNet-18 network. 
    Each method is evaluated the end of each task on the entire test set which includes all 100 classes. 
    \texttt{AdaLin} steadily outperforms ReLU, CReLU, and Deep Fourier, demonstrating robust performance 
    as the number of classes grows.}
    \label{fig:result_resnet_cifar100}
\end{figure}

Figure~\ref{fig:result_resnet_cifar100} presents our class-incremental CIFAR-100 results using a ResNet-18 backbone.
As training progresses and additional classes are introduced, test accuracy of each approach steadily improves, but \texttt{AdaLin}
consistently outperforms the other baselines. Notably, \texttt{AdaLin} achieves a higher final accuracy without the
abrupt performance drops typically seen in continual learning tasks. This outcome indicates that \texttt{AdaLin}’s
adaptive injection of linearity can effectively mitigate loss of plasticity, even in a large-scale, class-incremental
setting with a deep architecture like ResNet-18.

\subsection{Reinforcement Learning}

We conducted Reinforcement Learning (RL) experiments in an off-policy setting, which is known to be prone to loss of plasticity ~\citep{sokar2023dormant, galashov2025non}. Specifically, we trained a Soft Actor-Critic (SAC) ~\citep{haarnoja2018soft} agent across multiple MuJoCo environments, including Hopper-v4, Ant-v4, and HalfCheetah-v4. We use the default parameters provided by the CleanRL ~\citep{huang2022cleanrl} implementation and focus on investigating the impact of the replay ratio, which defines the number of gradient updates performed per environment interaction.

\begin{figure}[!t] % [h] means 'here' (preferred position)
    \centering \includegraphics[width=0.8\textwidth]{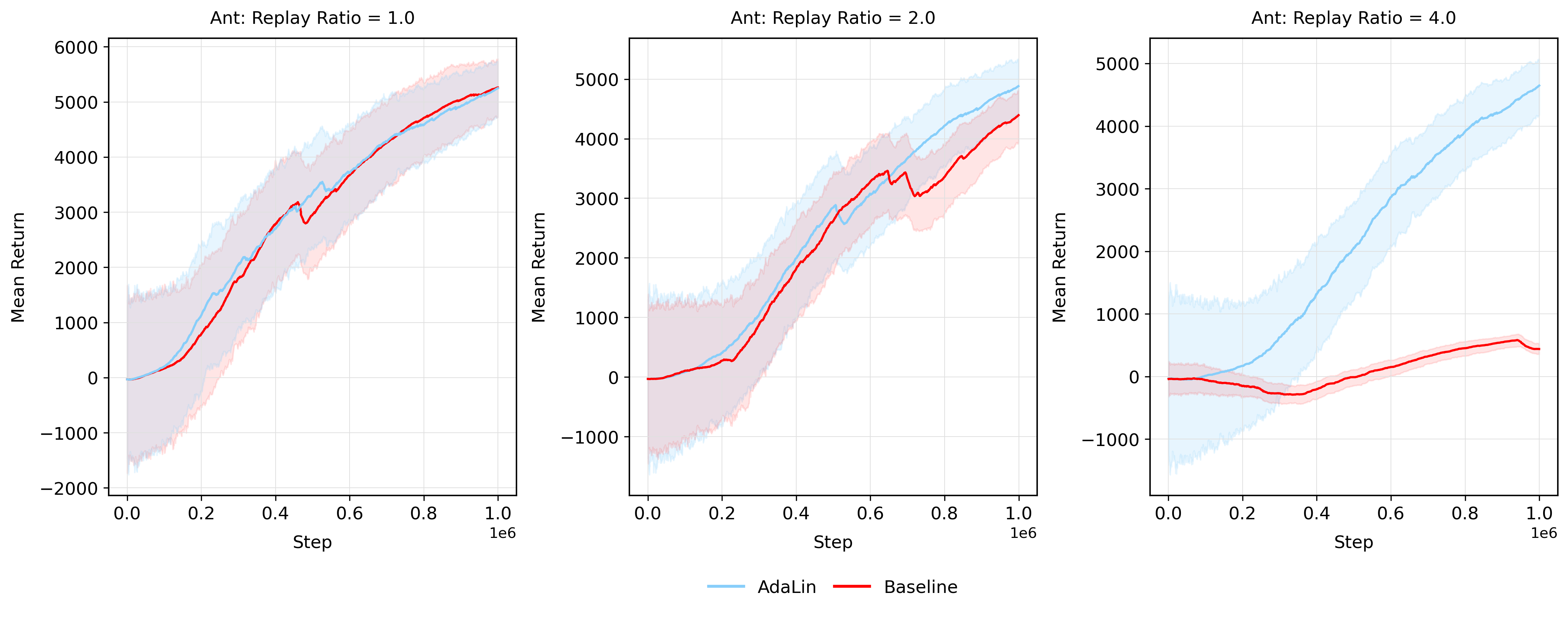}
    \caption{The AdaLin-enhanced model better maintains adaptability and mitigates plasticity loss under higher replay ratios in the MuJoCo environments. Performance comparison of the baseline Soft Actor-Critic (SAC) model using standard ReLU activation versus the AdaLin-enhanced SAC model across different replay ratios (1, 2, and 4). As the replay ratio increases, the baseline model struggles to learn, but the same architecture with AdaLin keeps its ability to adapt, leading to a notable performance gap.}

    \label{fig:result_RL}
\end{figure}

To systematically explore the effect of replay ratio, we evaluated three values: 1, 2, and 4. A higher replay ratio allows more gradient updates per environment interaction, thereby improving sample efficiency. However, it also increases the off-policyness of training, which can lead to overfitting to early transitions and, consequently, a loss of plasticity, which is studied in prior work ~\citet{kumar2020implicit}. As shown in Figure~\ref{fig:result_RL}, increasing the replay ratio results in a growing performance gap between the baseline SAC agent with the ReLU activation and the \texttt{AdaLin}-enhanced SAC model using ReLU as the base activation. This indicates that \texttt{AdaLin} better maintains adaptability under higher off-policy pressure. We obtain similar results for other MuJoCo environments and can be found in Appendix \ref{appendix:rl}

\begin{figure}[htb!] % [h] means 'here' (preferred position)
    \centering
    \includegraphics[width=0.7\textwidth]{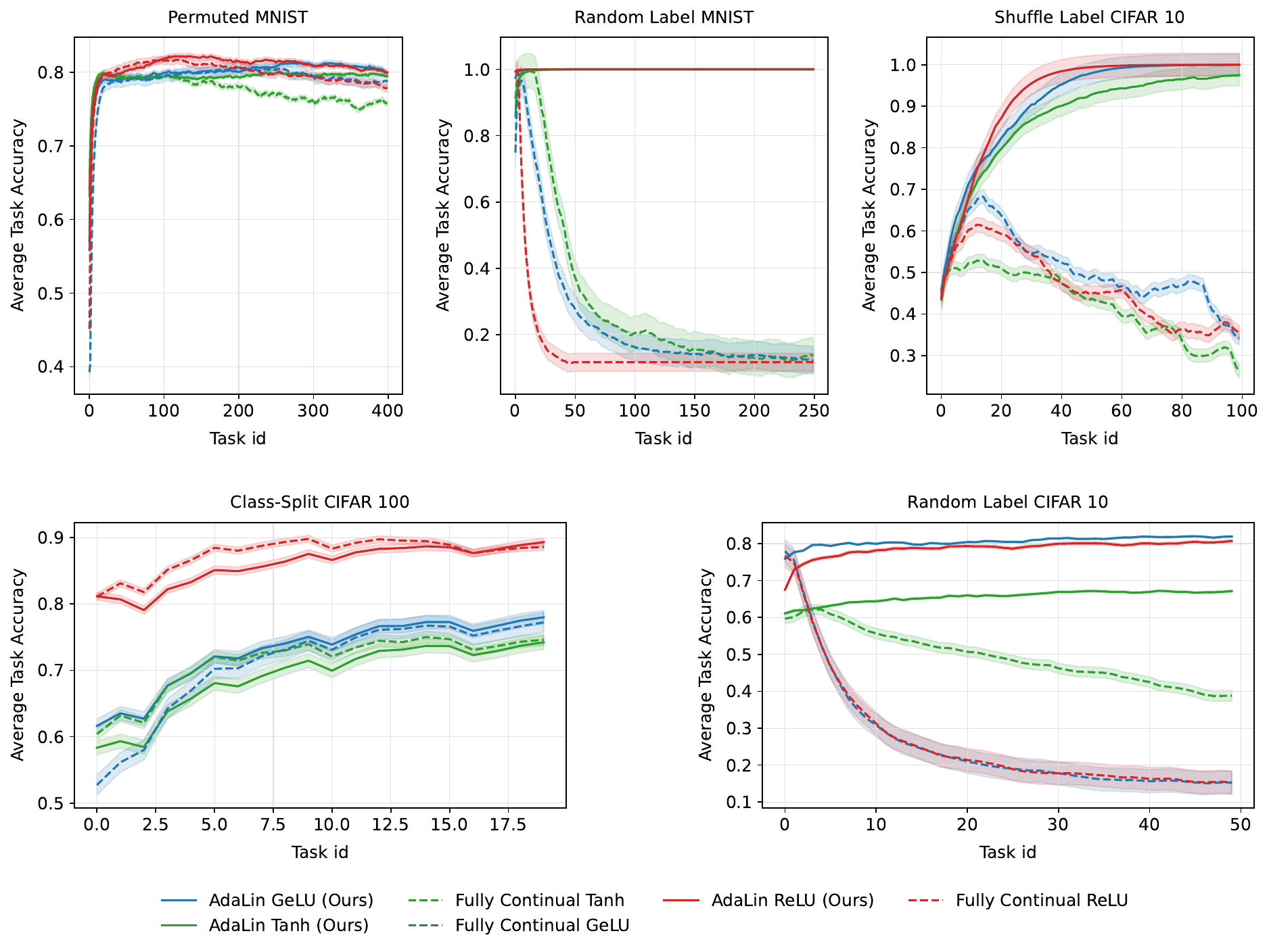}
    \caption{Performance comparison of different nonlinear activation functions with and without AdaLin. Without the adaptive linearity injection, activations such as ReLU, Tanh, and GELU exhibit severe plasticity loss over successive tasks, leading to a marked decline in average training accuracy. When used in conjunction with \texttt{AdaLin}, these activation functions demonstrate significantly improved performance, indicating that adaptive linearity injection effectively mitigates issues related to saturation and supports sustained gradient flow in continual learning scenarios.}
    \vspace{-3mm}
    \label{fig:other_activations}
\end{figure}

% \vspace{-2mm}
% \subsection{Ablation Studies}
% \vspace{-2mm}

\subsection{Using Different Activation Functions}
% \vspace{-2mm}

In this section, we empirically evaluate the effectiveness of our method on nonlinear activations beyond ReLU. Specifically, we consider Tanh and GELU ~\citep{hendrycks2016gaussian} as alternative base activation functions. As shown in Figure \ref{fig:other_activations}, both Tanh and GELU, when used continually, perform poorly across all benchmarks, exhibiting severe plasticity loss. However, using these activations in conjunction with \texttt{AdaLin} significantly improves the performance of the resulting network and successfully mitigates the loss of plasticity.   
% after incorporating the proposed module from Section \ref{sec:adalin}, 
% their performance improved significantly, and they no longer suffered from plasticity loss. 

This experiment highlights the generality of our approach: \textit{While \texttt{AdaLin} recovers PReLU when $\phi$ is ReLU, it effectively mitigates plasticity loss in cases where other activation functions originally struggled.}
\section{Discussion}\label{sec:conclusion}
% \vspace{-2mm}\

We introduced \texttt{AdaLin}, a novel approach that preserves plasticity in continual learning by dynamically injecting linearity into activation functions. Unlike previous methods that rely on fixed modifications or heuristic resets, \texttt{AdaLin} equips each neuron with a learnable injection parameter that adaptively balances linear and nonlinear components based on its saturation state. This mechanism sustains gradient flow in saturated regions and helps neurons maintain learning capacity over extended sequences of tasks.

Our experiments on benchmarks including Random Label MNIST, Random Label CIFAR-10, Permuted MNIST, and Class-Split CIFAR 100 demonstrate that \texttt{AdaLin} consistently mitigates the loss of plasticity observed in standard models. Ablation studies confirm that neuron-level adaptation is critical for preserving both trainability and generalization, while analysis of network dynamics reveals that \texttt{AdaLin} effectively maintains the gradient flow in the network by assigning small negative linear components to saturated neurons. Overall, \texttt{AdaLin} offers a practical solution for mitigating plasticity loss without extra hyperparameters or explicit task boundaries. 

More broadly, this work suggests that revisiting the basic building blocks of neural networks, such as activation functions, and endowing them with adaptive mechanisms may be important to enable successful continual learning. Future work will explore extending these ideas to more complex architectures, larger datasets, and integrating them with alternative strategies to combat catastrophic forgetting---moving closer to truly lifelong, adaptable learning systems.

\bibliography{collas2025_conference}

\begin{thebibliography}{31}
\providecommand{\natexlab}[1]{#1}
\providecommand{\url}[1]{\texttt{#1}}
\expandafter\ifx\csname urlstyle\endcsname\relax
  \providecommand{\doi}[1]{doi: #1}\else
  \providecommand{\doi}{doi: \begingroup \urlstyle{rm}\Url}\fi

\bibitem[Abbas et~al.(2023)Abbas, Zhao, Modayil, White, and Machado]{abbas2023loss}
Zaheer Abbas, Rosie Zhao, Joseph Modayil, Adam White, and Marlos~C Machado.
\newblock Loss of plasticity in continual deep reinforcement learning.
\newblock In \emph{Conference on lifelong learning agents}, pp.\  620--636. PMLR, 2023.

\bibitem[Apicella et~al.(2021)Apicella, Donnarumma, Isgr{\`o}, and Prevete]{apicella2021survey}
Andrea Apicella, Francesco Donnarumma, Francesco Isgr{\`o}, and Roberto Prevete.
\newblock A survey on modern trainable activation functions.
\newblock \emph{Neural Networks}, 138:\penalty0 14--32, 2021.

\bibitem[Ash \& Adams(2020)Ash and Adams]{ash2020warm}
Jordan Ash and Ryan~P Adams.
\newblock On warm-starting neural network training.
\newblock \emph{Advances in neural information processing systems}, 33:\penalty0 3884--3894, 2020.

\bibitem[Bittner et~al.(2015)Bittner, Milstein, and Grienberger]{Bittner2015}
Katherine~C. Bittner, Aaron~D. Milstein, and Christine Grienberger.
\newblock Behavioral time scale synaptic plasticity underlies ca1 place fields.
\newblock \emph{Nature Neuroscience}, 18:\penalty0 103--110, 2015.

\bibitem[Cheadle et~al.(2014)Cheadle, Wyart, and Tsetsos]{Cheadle2014}
Samuel Cheadle, Valentin Wyart, and Konstantinos Tsetsos.
\newblock Adaptive gain control during human perceptual decision-making.
\newblock \emph{Neuron}, 81:\penalty0 141--152, 2014.

\bibitem[Chung et~al.(2025)Chung, Cherif, Precup, and Meger]{chung2025parseval}
Wesley Chung, Lynn Cherif, Doina Precup, and David Meger.
\newblock Parseval regularization for continual reinforcement learning.
\newblock \emph{Advances in Neural Information Processing Systems}, 37:\penalty0 127937--127967, 2025.

\bibitem[Dohare et~al.(2024)Dohare, Hernandez-Garcia, Lan, Rahman, Mahmood, and Sutton]{dohare2024loss}
Shibhansh Dohare, J~Fernando Hernandez-Garcia, Qingfeng Lan, Parash Rahman, A~Rupam Mahmood, and Richard~S Sutton.
\newblock Loss of plasticity in deep continual learning.
\newblock \emph{Nature}, 632\penalty0 (8026):\penalty0 768--774, 2024.

\bibitem[Ferguson \& Cardin(2020)Ferguson and Cardin]{Ferguson2020}
Katie~A. Ferguson and Jessica~A. Cardin.
\newblock Gain control by interneurons in the neocortex.
\newblock \emph{Nature Reviews Neuroscience}, 21:\penalty0 133--142, 2020.

\bibitem[Galashov et~al.(2025)Galashov, Titsias, Gy{\"o}rgy, Lyle, Pascanu, Teh, and Sahani]{galashov2025non}
Alexandre Galashov, Michalis Titsias, Andr{\'a}s Gy{\"o}rgy, Clare Lyle, Razvan Pascanu, Yee~Whye Teh, and Maneesh Sahani.
\newblock Non-stationary learning of neural networks with automatic soft parameter reset.
\newblock \emph{Advances in Neural Information Processing Systems}, 37:\penalty0 83197--83234, 2025.

\bibitem[Gambino \& Holtmaat(2014)Gambino and Holtmaat]{Gambino2014}
Fabrizio Gambino and Anthony Holtmaat.
\newblock Dendritic nmda spikes are necessary for associative long-term potentiation in vivo.
\newblock \emph{Nature}, 515:\penalty0 116--119, 2014.

\bibitem[Gidon et~al.(2020)Gidon, Zolnik, and Fidzinski]{Gidon2020}
Albert Gidon, Taylor~A. Zolnik, and Pawel Fidzinski.
\newblock Dendritic action potentials and computation in human layer 2/3 cortical neurons.
\newblock \emph{Science}, 367:\penalty0 83--87, 2020.

\bibitem[Haarnoja et~al.(2018)Haarnoja, Zhou, Abbeel, and Levine]{haarnoja2018soft}
Tuomas Haarnoja, Aurick Zhou, Pieter Abbeel, and Sergey Levine.
\newblock Soft actor-critic: Off-policy maximum entropy deep reinforcement learning with a stochastic actor.
\newblock In \emph{International conference on machine learning}, pp.\  1861--1870. Pmlr, 2018.

\bibitem[He et~al.(2015)He, Zhang, Ren, and Sun]{he2015delving}
Kaiming He, Xiangyu Zhang, Shaoqing Ren, and Jian Sun.
\newblock Delving deep into rectifiers: Surpassing human-level performance on imagenet classification.
\newblock In \emph{Proceedings of the IEEE international conference on computer vision}, pp.\  1026--1034, 2015.

\bibitem[Hendrycks \& Gimpel(2016)Hendrycks and Gimpel]{hendrycks2016gaussian}
Dan Hendrycks and Kevin Gimpel.
\newblock Gaussian error linear units (gelus).
\newblock \emph{arXiv preprint arXiv:1606.08415}, 2016.

\bibitem[Huang et~al.(2022)Huang, Dossa, Ye, Braga, Chakraborty, Mehta, and Araújo]{huang2022cleanrl}
Shengyi Huang, Rousslan Fernand~Julien Dossa, Chang Ye, Jeff Braga, Dipam Chakraborty, Kinal Mehta, and João~G.M. Araújo.
\newblock Cleanrl: High-quality single-file implementations of deep reinforcement learning algorithms.
\newblock \emph{Journal of Machine Learning Research}, 23\penalty0 (274):\penalty0 1--18, 2022.
\newblock URL \url{http://jmlr.org/papers/v23/21-1342.html}.

\bibitem[Kumar et~al.(2020)Kumar, Agarwal, Ghosh, and Levine]{kumar2020implicit}
Aviral Kumar, Rishabh Agarwal, Dibya Ghosh, and Sergey Levine.
\newblock Implicit under-parameterization inhibits data-efficient deep reinforcement learning.
\newblock \emph{arXiv preprint arXiv:2010.14498}, 2020.

\bibitem[Kumar et~al.(2023)Kumar, Marklund, and Van~Roy]{kumar2023maintaining}
Saurabh Kumar, Henrik Marklund, and Benjamin Van~Roy.
\newblock Maintaining plasticity in continual learning via regenerative regularization.
\newblock \emph{arXiv preprint arXiv:2308.11958}, 2023.

\bibitem[Lee et~al.(2024)Lee, Cho, Kim, Kim, Min, Choo, and Lyle]{lee2024slow}
Hojoon Lee, Hyeonseo Cho, Hyunseung Kim, Donghu Kim, Dugki Min, Jaegul Choo, and Clare Lyle.
\newblock Slow and steady wins the race: Maintaining plasticity with hare and tortoise networks.
\newblock \emph{arXiv preprint arXiv:2406.02596}, 2024.

\bibitem[Lewandowski et~al.(2023)Lewandowski, Tanaka, Schuurmans, and Machado]{lewandowski2023curvature}
Alex Lewandowski, Haruto Tanaka, Dale Schuurmans, and Marlos~C Machado.
\newblock Curvature explains loss of plasticity.
\newblock 2023.

\bibitem[Lewandowski et~al.(2024)Lewandowski, Schuurmans, and Machado]{lewandowski2024plastic}
Alex Lewandowski, Dale Schuurmans, and Marlos~C Machado.
\newblock Plastic learning with deep fourier features.
\newblock \emph{arXiv preprint arXiv:2410.20634}, 2024.

\bibitem[Lovett-Barron \& Losonczy(2014)Lovett-Barron and Losonczy]{Lovett2014}
Matt Lovett-Barron and Attila Losonczy.
\newblock Dendritic inhibition in the hippocampus supports fear learning.
\newblock \emph{Science}, 343:\penalty0 857--860, 2014.

\bibitem[Lyle et~al.(2023)Lyle, Zheng, Nikishin, Pires, Pascanu, and Dabney]{lyle2023understanding}
Clare Lyle, Zeyu Zheng, Evgenii Nikishin, Bernardo~Avila Pires, Razvan Pascanu, and Will Dabney.
\newblock Understanding plasticity in neural networks.
\newblock In \emph{International Conference on Machine Learning}, pp.\  23190--23211. PMLR, 2023.

\bibitem[Lyle et~al.(2024)Lyle, Zheng, Khetarpal, van Hasselt, Pascanu, Martens, and Dabney]{lyle2024disentangling}
Clare Lyle, Zeyu Zheng, Khimya Khetarpal, Hado van Hasselt, Razvan Pascanu, James Martens, and Will Dabney.
\newblock Disentangling the causes of plasticity loss in neural networks.
\newblock \emph{arXiv preprint arXiv:2402.18762}, 2024.

\bibitem[McGinley et~al.(2015)McGinley, David, and McCormick]{McGinley2015}
Matthew~J. McGinley, Stephen~V. David, and David~A. McCormick.
\newblock Cortical arousal modulates sensory processing in mouse visual cortex.
\newblock \emph{Neuron}, 87:\penalty0 323--331, 2015.

\bibitem[Misra(2019)]{misra2019mish}
Diganta Misra.
\newblock Mish: A self regularized non-monotonic activation function.
\newblock \emph{arXiv preprint arXiv:1908.08681}, 2019.

\bibitem[Nikishin et~al.(2023)Nikishin, Oh, Ostrovski, Lyle, Pascanu, Dabney, and Barreto]{nikishin2023deep}
Evgenii Nikishin, Junhyuk Oh, Georg Ostrovski, Clare Lyle, Razvan Pascanu, Will Dabney, and Andr{\'e} Barreto.
\newblock Deep reinforcement learning with plasticity injection.
\newblock \emph{Advances in Neural Information Processing Systems}, 36:\penalty0 37142--37159, 2023.

\bibitem[Palacios \& Gambino(2021)Palacios and Gambino]{Palacios2021}
Andrea~G. Palacios and Fabrizio Gambino.
\newblock Experience-dependent plasticity of dendritic plateau potentials in sensory cortex.
\newblock \emph{PNAS}, 118:\penalty0 e2012510118, 2021.

\bibitem[Qiu et~al.(2018)Qiu, Xu, and Cai]{qiu2018frelu}
Suo Qiu, Xiangmin Xu, and Bolun Cai.
\newblock Frelu: Flexible rectified linear units for improving convolutional neural networks.
\newblock In \emph{2018 24th international conference on pattern recognition (icpr)}, pp.\  1223--1228. IEEE, 2018.

\bibitem[Rao \& Geffen(2017)Rao and Geffen]{Rao2017}
Rajesh~P.N. Rao and Maria~N. Geffen.
\newblock Inhibitory interneurons gate adaptive sensory processing in the auditory cortex.
\newblock \emph{Cell Reports}, 16:\penalty0 1060--1067, 2017.

\bibitem[Sokar et~al.(2023)Sokar, Agarwal, Castro, and Evci]{sokar2023dormant}
Ghada Sokar, Rishabh Agarwal, Pablo~Samuel Castro, and Utku Evci.
\newblock The dormant neuron phenomenon in deep reinforcement learning.
\newblock In \emph{International Conference on Machine Learning}, pp.\  32145--32168. PMLR, 2023.

\bibitem[Trottier et~al.(2017)Trottier, Giguere, and Chaib-Draa]{trottier2017parametric}
Ludovic Trottier, Philippe Giguere, and Brahim Chaib-Draa.
\newblock Parametric exponential linear unit for deep convolutional neural networks.
\newblock In \emph{2017 16th IEEE international conference on machine learning and applications (ICMLA)}, pp.\  207--214. IEEE, 2017.

\end{thebibliography}
\bibliographystyle{collas2025_conference}
\newpage
\appendix\label{sec:appendix}
\section{Benchmarks}
\label{appendix:benchmarks}
Our experimental framework comprises five tasks that cover these scenarios:
\begin{itemize}
    \item \textbf{Permuted MNIST.} For this setting, we randomly select 10,000 images from the MNIST dataset. A unique, fixed random permutation is then applied to the pixel order of all images in each task. This permutation scrambles the original spatial arrangement of the pixels, effectively creating a new input distribution for every task. Although the digit labels remain unchanged, the input representation is entirely different from the original, requiring the model to relearn the mapping between the permuted inputs and the correct digit labels. This task is designed to probe the network’s capacity to adapt to a severe change in the input structure while still retaining its ability to classify accurately.  We perform this experiment on 400 tasks, and importantly, in each task, the data is only seen once (1 epoch) with a batch size of 16, making the learning process more challenging by limiting the model’s exposure to the transformed data.

    \item \textbf{Random Label MNIST.}
    We randomly sample 1,200 images from the MNIST dataset and assign completely random labels to these images in each task, discarding the true digit labels. In every task, the inherent structure linking the image to its correct label is removed, and the network is forced to memorize an arbitrary association between each image and its new random label. This scenario stresses the network’s memorization capacity, allowing us to study how overfitting to noise impacts plasticity—specifically, how the ability to adapt to new tasks might deteriorate when the network has learned to memorize unstructured, random targets. We run this experiment for 250 tasks, training the network for 200 epochs per task with a batch size of 16, ensuring that the model undergoes extensive exposure to the randomized associations.
    \item \textbf{Random-Label CIFAR 10.}
    Similar in principle to Random Label MNIST, this setting uses 1,200 randomly selected images from the CIFAR 10 dataset, with each image assigned a label drawn at random from the available classes. Given that CIFAR 10 images are more visually complex and colorful than MNIST digits, this task challenges the network to memorize more intricate features without any meaningful mapping to the correct class. By evaluating the network’s performance on this task, we can assess how the memorization of arbitrary targets affects plasticity, especially in scenarios where the visual information is richer and the noise in the labels is more pronounced. To ensure the model has ample time to fully internalize these complex, unstructured associations, we train it for a substantial 700 epochs per task. This extended training regimen, applied across 100 tasks with a batch size of 16, allows us to thoroughly examine the long-term effects of memorization on plasticity.  

    \item \textbf{Random Shuffle CIFAR 10.}
    In each task, we use 5,000 samples from the CIFAR 10 dataset. Unlike the random-label tasks, here the images in the same class will have the same label, but the mapping between the images and their class labels is randomly shuffled for each task. This setup introduces a controlled, slight distribution shift: while the underlying image content does not change, the correspondence between images and class labels is systematically altered. The goal is to analyze how subtle changes in label associations impact the network’s classification ability and its adaptability to systematic shifts in the data distribution. This task helps to isolate the effect of class-level label reassignments on plasticity. 
    The rotative change in the class distribution of each task can be inherently considered as an online learning setting, requiring the model to continuously update its parameters in response to the reassignment of the class labels. To capture the effects, we run this experiment for 100 tasks, training the model for 20 epochs per task with a batch size of 16, ensuring that while each task is learned relatively quickly, the cumulative impact of repeated shifts can be observed over time.
    \item \textbf{Class-Split CIFAR 100} In our Class-Split CIFAR 100 task, data is drawn from the CIFAR 100 dataset to create a realistic continual learning scenario. In each task, the model is presented with image–label pairs from 5 new classes, with 500 samples provided per class, resulting in a total of 2,500 data pairs per task. The model is trained for 20 epochs per task with a batch size of 32, proceeding until all 100 classes have been introduced. This environment is designed to simulate scenarios with shifting input distributions, where the network must rapidly adapt to new classes. The task thereby serves as a rigorous benchmark for evaluating methods aimed at mitigating plasticity loss.
    
\end{itemize}

\section{Training Setup}\label{sec:train_setup}
\subsection{Architecture}
Similar to previous work~\citep{kumar2023maintaining} we adopt two types of network architectures in our experiments: a multi-layer perceptron (MLP) and a convolutional
neural network (CNN). To be able to analyze the behavior of the network on the neuron level, we use the same MLP for the Permuted MNIST, Random Label
MNIST, Shuffle Label CIFAR-10, and Random Label CIFAR-10 tasks, and reserve the CNN for the Continual
CIFAR-100 task, where the input complexity necessitates a deeper architecture. For the architectures of these two networks, we use the ones applied in ~\cite{kumar2023maintaining}:

\paragraph{MLP.}
We employ a two-layer MLP with 100 hidden units per layer. Each hidden layer is followed by the activation function
under study (e.g., ReLU, Tanh, or our proposed AdaLin). the final output layer consists of 10 units, accordingly. This design provides sufficient capacity to study
plasticity loss and neuron behavior under various forms of non-stationarity without overwhelming computational resources.

\paragraph{CNN.}
For the Continual CIFAR-100 task, we use a two-layer convolutional network, each convolutional layer having a
kernel size of $5 \times 5$ with 16 output channels, followed by a max-pooling operation. The final layers are fully
connected, with the activation function placed in between. The first fully connected layer maps from $16 \times 5 \times
5$ down to 64 units, and the second fully connected layer outputs 100 units (one per CIFAR-100 class). This CNN
structure is designed to capture the higher complexity of CIFAR-100 images while still allowing us to isolate and study
the effects of plasticity loss in a more realistic, larger-scale setting.

{\paragraph{Resnet-18.}
We employed ResNet-18 to assess how effectively AdaLin integrates with contemporary deep architectures in large-scale models. To suit the dataset’s smaller image dimensions, we removed the stem layers, and we implemented a gradient clipping threshold of 0.5 to maintain training stability.
\subsection{Hyperparameters}
We employed the SGD optimizer for all models in our experiments. A hyperparameter sweep was performed over 3 random seeds for each baseline–benchmark pair, sweeping over learning rates \(\{1e^{-2},\,1e^{-3},\,1e^{-4}\}\) and batch sizes \(\{16,\,32,\,64\}\), selecting the configuration that maximized the total average online accuracy across these seeds. Once the best hyperparameters were identified, each experiment was run on three additional seeds, and we report the mean outcome, with shaded regions in our plots representing one standard deviation. All reported metrics (e.g., average task accuracy) are computed based on these final runs. For Permuted MNIST, we followed the common practice of training for only one epoch per task~\citep{dohare2024loss,kumar2023maintaining}, while for the other benchmarks we used a sufficient number of epochs to ensure that each model achieved a reasonable accuracy on the first task. This setup balances comparability with prior work and ensures that our comparisons reflect robust trends rather than artifacts of insufficient training or overly favorable initial conditions. We report the hyperparameters used for each plasticity benchmark in Tables \ref{tab:benchmark1_hyperparams} to \ref{tab:benchmark5_hyperparams}. Additionally, Table \ref{tab:times} presents the training systems, GPUs used, and training durations for various machine learning tasks conducted during the experiment.

% \begin{table}[h]
%     \centering
%     \caption{Model Architecture and Optimal Hyper-parameters on \textbf{Permuted Mnist}}
%     \label{tab:benchmark1_hyperparams}
%     \begin{tabular}{l l l p{5cm}}
%         \toprule
%         \textbf{Method} & \textbf{Optimizer} & \textbf{Architecture} & \textbf{Optimal Hyper-parameters} \\
%         \midrule
%         Baseline        & SGD    & MLP  & lr $= 1\mathrm{e}{-2}$, batch = 16, epochs = 1 \\
%          \texttt{AdaLin}        & SGD    & MLP  & lr $= 1\mathrm{e}{-2}$, batch = 16, epochs = 1 \\
%          Deep Fourier        & SGD    & MLP  & lr $= 1\mathrm{e}{-2}$, batch = 16, epochs = 1 \\CReLU        & SGD    & MLP  & lr $= 1\mathrm{e}{-2}$, batch = 16, epochs = 1 \\
%          Deep Linear        & SGD    & MLP  & lr $= 1\mathrm{e}{-2}$, batch = 16, epochs = 1 \\
%          Scratch        & SGD    & MLP  & lr $= 1\mathrm{e}{-2}$, batch = 16, epochs = 1 \\
%          L2        & SGD    & MLP  & lr $= 1\mathrm{e}{-2}$, batch = 16, epochs = 1, $\lambda = 0.01$ \\
%          Shrink \& Perturb        & SGD    & MLP  & lr $= 1\mathrm{e}{-2}$, batch = 16, epochs = 1, $p = 1 - 1\mathrm{e}{-4}$, $\sigma = 1\mathrm{e}{-2}$ \\
%         \bottomrule
%     \end{tabular}
% \end{table}

\begin{table}[h]
    \centering
    \caption{Model Architecture and Optimal Hyperparameters on \textbf{Permuted Mnist}}
    \label{tab:benchmark1_hyperparams}
    \begin{tabular}{l l p{5cm}}
        \toprule
        \textbf{Method} & \textbf{Architecture} & \textbf{Optimal Hyperparameters} \\
        \midrule
        \texttt{Baseline}              & MLP  & lr $= 1\mathrm{e}{-2}$, batch = 16, epochs = 1 \\
        \texttt{AdaLin}       & MLP  & lr $= 1\mathrm{e}{-2}$, batch = 16, epochs = 1 \\
        \texttt{Deep Fourier}          & MLP  & lr $= 1\mathrm{e}{-2}$, batch = 16, epochs = 1 \\
        \texttt{CReLU}                 & MLP  & lr $= 1\mathrm{e}{-2}$, batch = 16, epochs = 1 \\
        \texttt{Deep Linear}           & MLP  & lr $= 1\mathrm{e}{-2}$, batch = 16, epochs = 1 \\
        \texttt{Scratch}               & MLP  & lr $= 1\mathrm{e}{-2}$, batch = 16, epochs = 1 \\
        \texttt{L2}                    & MLP  & lr $= 1\mathrm{e}{-2}$, batch = 16, epochs = 1, $\lambda = 0.01$ \\
        \texttt{Shrink and Perturb}     & MLP  & lr $= 1\mathrm{e}{-2}$, batch = 16, epochs = 1, $p = 1 - 1\mathrm{e}{-4}$, $\sigma = 1\mathrm{e}{-2}$ \\
        \bottomrule
    \end{tabular}
\end{table}

%===========================================================
% Table for Benchmark 2 (e.g., Random Label MNIST)
%===========================================================
\begin{table}[h]
    \centering
    \caption{Model Architecture and Optimal Hyperparameters on \textbf{Random Label MNIST}}
    \label{tab:benchmark2_hyperparams}
    \begin{tabular}{l l p{5cm}}
        \toprule
        \textbf{Method} & \textbf{Architecture} & \textbf{Optimal Hyperparameters} \\
        \midrule
        \texttt{Baseline}              & MLP  & lr $= 1\mathrm{e}{-2}$, batch = 16, epochs = 200 \\
        \texttt{AdaLin}       & MLP  & lr $= 1\mathrm{e}{-2}$, batch = 16, epochs = 200 \\
        \texttt{Deep Fourier}          & MLP  & lr $= 1\mathrm{e}{-2}$, batch = 16, epochs = 200 \\
        \texttt{CReLU}                 & MLP  & lr $= 1\mathrm{e}{-2}$, batch = 16, epochs = 200 \\
        \texttt{Deep Linear}           & MLP  & lr $= 1\mathrm{e}{-2}$, batch = 16, epochs = 200 \\
        \texttt{Scratch}               & MLP  & lr $= 1\mathrm{e}{-2}$, batch = 16, epochs = 200 \\
        \texttt{L2}                    & MLP  & lr $= 1\mathrm{e}{-2}$, batch = 16, epochs = 200, $\lambda = 0.01$ \\
        \texttt{Shrink and Perturb}     & MLP  & lr $= 1\mathrm{e}{-2}$, batch = 16, epochs = 200, $p = 1 - 1\mathrm{e}{-4}$, $\sigma = 1\mathrm{e}{-2}$ \\
        \bottomrule
    \end{tabular}
\end{table}

%===========================================================
% Table for Benchmark 3
%===========================================================
\begin{table}[htb]
    \centering
    \caption{Model Architecture and Optimal Hyperparameters on \textbf{Shuffel Label Cifar 10}}
    \label{tab:benchmark3_hyperparams}
    \begin{tabular}{l l p{5cm}}
        \toprule
        \textbf{Method} & \textbf{Architecture} & \textbf{Optimal Hyperparameters} \\
        \midrule
        \texttt{Baseline}              & MLP  & lr $= 1\mathrm{e}{-2}$, batch = 16, epochs = 20 \\
        \texttt{AdaLin}       & MLP  & lr $= 1\mathrm{e}{-2}$, batch = 16, epochs = 20 \\
        \texttt{Deep Fourier}          & MLP  & lr $= 1\mathrm{e}{-2}$, batch = 16, epochs = 20 \\
        \texttt{CReLU}                 & MLP  & lr $= 1\mathrm{e}{-2}$, batch = 16, epochs = 20 \\
        \texttt{Deep Linear}           & MLP  & lr $= 1\mathrm{e}{-2}$, batch = 16, epochs = 20 \\
        \texttt{Scratch}               & MLP  & lr $= 1\mathrm{e}{-2}$, batch = 16, epochs = 20 \\
        \texttt{L2}                    & MLP  & lr $= 1\mathrm{e}{-2}$, batch = 16, epochs = 20, $\lambda = 0.01$ \\
        \texttt{Shrink and Perturb}     & MLP  & lr $= 1\mathrm{e}{-2}$, batch = 16, epochs = 20, $p = 1 - 1\mathrm{e}{-4}$, $\sigma = 1\mathrm{e}{-2}$ \\
        \bottomrule
    \end{tabular}
\end{table}

%===========================================================
% Table for Benchmark 4
%===========================================================
\begin{table}[htb]
    \centering
    \caption{Model Architecture and Optimal Hyperparameters on \textbf{Random Label CIFAR 10}}
    \label{tab:benchmark4_hyperparams}
    \begin{tabular}{l l p{5cm}}
        \toprule
        \textbf{Method} & \textbf{Architecture} & \textbf{Optimal Hyperparameters} \\
        \midrule
        \texttt{Baseline}              & MLP  & lr $= 1\mathrm{e}{-2}$, batch = 16, epochs = 700 \\
        \texttt{AdaLin}       & MLP  & lr $= 1\mathrm{e}{-2}$, batch = 16, epochs = 700 \\
        \texttt{Deep Fourier}          & MLP  & lr $= 1\mathrm{e}{-3}$, batch = 32, epochs = 700 \\
        \texttt{CReLU}                 & MLP  & lr $= 1\mathrm{e}{-2}$, batch = 16, epochs = 700 \\
        \texttt{Deep Linear}           & MLP  & lr $= 1\mathrm{e}{-2}$, batch = 16, epochs = 700 \\
        \texttt{Scratch}               & MLP  & lr $= 1\mathrm{e}{-2}$, batch = 16, epochs = 700 \\
        \texttt{L2}                    & MLP  & lr $= 1\mathrm{e}{-2}$, batch = 16, epochs = 700, $\lambda = 0.01$ \\
        \texttt{Shrink and Perturb}     & MLP  & lr $= 1\mathrm{e}{-2}$, batch = 16, epochs = 700, $p = 1 - 1\mathrm{e}{-4}$, $\sigma = 1\mathrm{e}{-2}$ \\
        \bottomrule
    \end{tabular}
\end{table}

%===========================================================
% Table for Benchmark 5
%===========================================================
\begin{table}[htb]
    \centering
    \caption{Model Architecture and Optimal Hyperparameters on \textbf{Class-Split CIFAR 100}}
    \label{tab:benchmark5_hyperparams}
    \begin{tabular}{l l p{5cm}}
        \toprule
        \textbf{Method} & \textbf{Architecture} & \textbf{Optimal Hyperparameters} \\
        \midrule
        \texttt{Baseline}              & CNN  & lr $= 1\mathrm{e}{-2}$, batch = 16, epochs = 20 \\
        \texttt{AdaLin}       & CNN  & lr $= 1\mathrm{e}{-2}$, batch = 16, epochs = 20 \\
        \texttt{Deep Fourier}          & CNN  & lr $= 1\mathrm{e}{-2}$, batch = 16, epochs = 20 \\
        \texttt{CReLU}                 & CNN  & lr $= 1\mathrm{e}{-2}$, batch = 16, epochs = 20 \\
        \texttt{Deep Linear}          & CNN  & lr $= 1\mathrm{e}{-2}$, batch = 16, epochs = 20 \\
        \texttt{Scratch}               & CNN  & lr $= 1\mathrm{e}{-2}$, batch = 16, epochs = 20 \\
        \texttt{L2}                    & CNN  & lr $= 1\mathrm{e}{-2}$, batch = 16, epochs = 20, $\lambda = 0.01$ \\
        \texttt{Shrink and Perturb}     & CNN  & lr $= 1\mathrm{e}{-2}$, batch = 16, epochs = 20, $p = 1 - 1\mathrm{e}{-4}$, $\sigma = 1\mathrm{e}{-2}$ \\
        \bottomrule
    \end{tabular}
\end{table}

%===========================================================
% Table for Benchmark 5
%===========================================================
\begin{table}[htb]
    \centering
    \caption{Model Architecture and Optimal Hyperparameters on \textbf{Class-Incremental CIFAR 100}}
    \label{tab:benchmark6_hyperparams}
    \begin{tabular}{l l p{5cm}}
        \toprule
        \textbf{Method} & \textbf{Architecture} & \textbf{Optimal Hyperparameters} \\
        \midrule
        \texttt{Baseline}              & Resnet  & lr $= 1\mathrm{e}{-3}$, batch = 250, time steps = 6000 \\
        \texttt{AdaLin}       & Resnet  & lr $= 1\mathrm{e}{-3}$, batch = 250, time steps = 6000 \\
        \texttt{Deep Fourier}          & Resnet  & lr $= 1\mathrm{e}{-3}$, batch = 250, time steps = 6000 \\
        \texttt{CReLU}                 & Resnet  & lr $= 1\mathrm{e}{-3}$, batch = 250, time steps = 6000 \\
        \bottomrule
    \end{tabular}
\end{table}

\begin{table}[H] % The [H] forces the table to appear here
\centering
\caption{Benchmarks, Used GPUs, and Training Time}
\label{tab:times} 
\begin{tabular}{|l|l|l|}
\hline
\textbf{Task} & \textbf{GPU Used} & \textbf{Time} \\ \hline
Permuted MNIST & RTX 3080 & 1 hour \\ \hline
Random Label MNIST & RTX 3080 & 1:30 hours \\ \hline
Shuffle Label CIFAR 10 & L40S & 2:30 hours \\ \hline
Random Label CIFAR 10 & L40S & 6 hours \\ \hline
Class-Split CIFAR 100 & RTX 3080 & 1 hour \\ \hline
\end{tabular}
\end{table}

\section{Baseline Methods}\label{sec:baselines}

In this section, we briefly describe the baseline methods used in our experiments:

\begin{itemize}
    
    \item \textbf{CReLU:} \citet{abbas2023loss} highlighted how standard ReLU activations contribute to activation collapse, leading to diminishing gradients and reduced learning capability in continual reinforcement learning. To counteract this, they introduced Concatenated ReLUs (CReLUs), which maintain activation diversity by duplicating activations with opposite signs, effectively ensuring that some portion of each neuron's output stays unsaturated, but it effectively doubles the number of outputs and parameters for each layer..  Their work underscores the critical role of activation function design in ensuring continual learning effectiveness in non-stationary environments.

    \item \textbf{Deep Fourier Features:}In \cite{lewandowski2024plastic}, the authors demonstrate that linear networks are resilient to plasticity loss, which has inspired the development of deep Fourier features. In this approach, each neuron's preactivation \(z\) is transformed into a two-dimensional feature by applying fixed sinusoidal functions:
\[
f(z) = \begin{bmatrix} \sin(z) \\ \cos(z) \end{bmatrix}.
\]
This dual-channel representation ensures that at any input, at least one of the outputs remains near-linear. To see this mathematically, note that the derivative of the cosine function is 
\[
\frac{d}{dz}\cos(z) = -\sin(z).
\]
Thus, whenever \(\sin(z)=0\) (i.e., when \(z = n\pi\) for some integer \(n\)), the derivative of \(\cos(z)\) vanishes. However, at these points the sine function has its maximum derivative (\(\frac{d}{dz}\sin(z)=\cos(z)=\pm1\)), ensuring that the overall representation retains a strong linear behavior. This complementary behavior guarantees robust gradient flow even in regions where one of the outputs saturates, effectively embedding a linear model within the nonlinear framework.

    \item \textbf{Deep Linear Network:} A network using only linear activations avoids plasticity loss entirely. However, its lack of nonlinearity severely limits expressivity, resulting in lower overall performance on complex tasks.
    
    \item \textbf{L2 Regularization:} This baseline applies an \(L_2\) penalty on the model's weights during training to constrain their magnitude. While this can help preserve plasticity, it may also restrict the model's capacity to learn rich representations.
    
    \item \textbf{Shrink \& Perturb:}  Shrink \& Perturb (S\&P)~\citep{ash2020warm} aims to preserve plasticity by jointly shrinking weights to reduce their magnitude and perturbing them with noise to encourage exploration. Inspired by warm-start methods, S\&P uses a single hyperparameter to govern both the intensity of the weight shrinkage and the level of added noise. As the weights converge, this procedure continually reintroduces variation into the parameter space, preventing the network from becoming overly rigid and thus helping maintain adaptability.

\end{itemize}

\section{Reinforcement learning}\label{appendix:rl}

In this section, we evaluate the performance of AdaLin in a reinforcement learning setting by training a Soft Actor-Critic (SAC) agent with varying replay ratios. We report the mean return for both the baseline model using ReLU and the AdaLin-enhanced model. As shown in Figure \ref{appendix:result_RL}, the performance gap between ReLU and AdaLin increases with the replay ratio, and this effect becomes more pronounced in tasks with higher-dimensional action spaces.

\begin{figure}[!t] % [h] means 'here' (preferred position)
    \centering \includegraphics[width=0.8\textwidth]{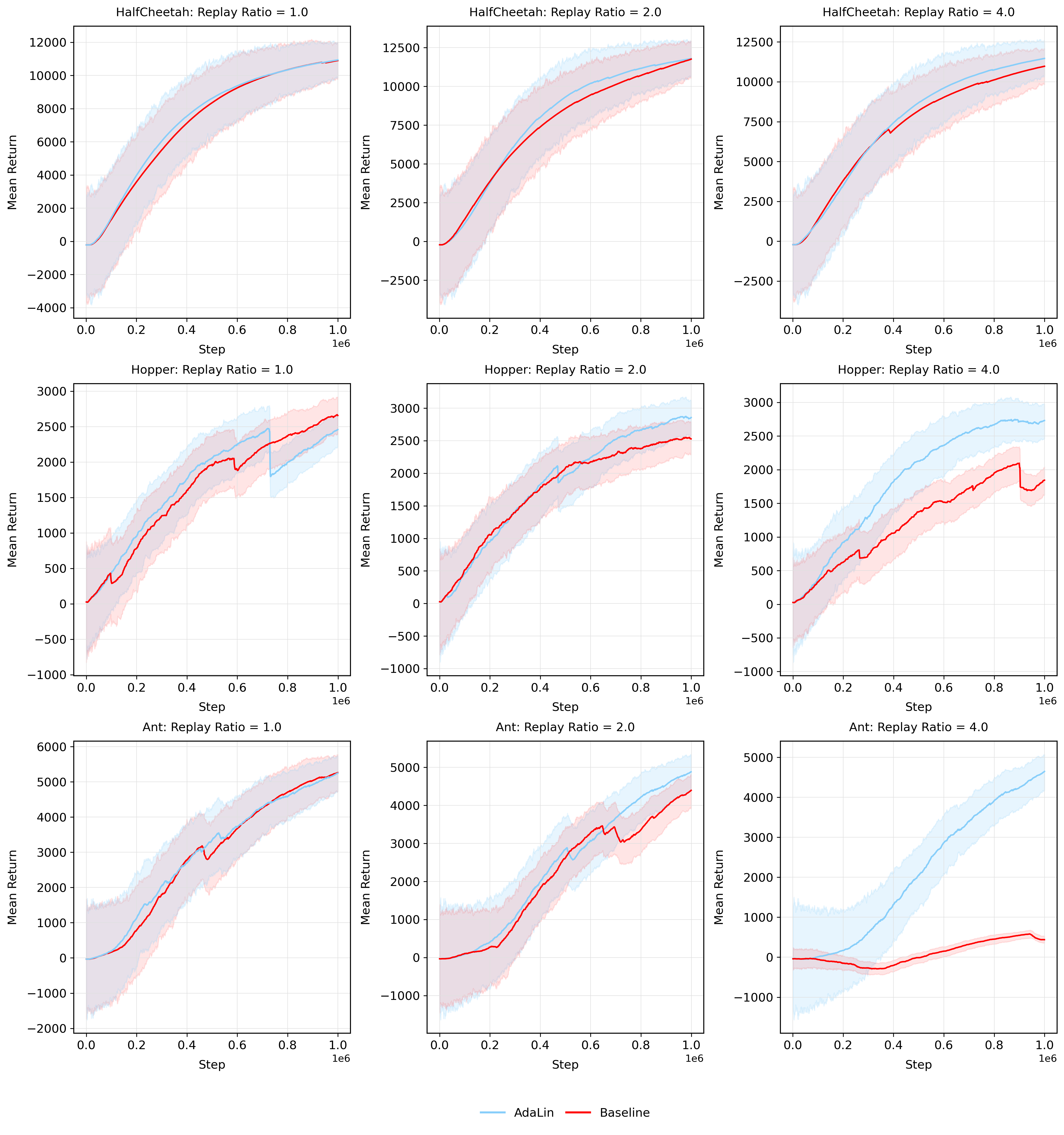}
    \caption{The AdaLin-enhanced model better maintains adaptability and mitigates plasticity loss under higher replay ratios in the MuJoCo environments. Performance comparison of the baseline Soft Actor-Critic (SAC) model using standard ReLU activation versus the AdaLin-enhanced SAC model across different replay ratios (1, 2, and 4). As the replay ratio increases, the baseline model struggles to learn, but the same architecture with AdaLin keeps its ability to adapt, leading to a notable performance gap.}

    \label{appendix:result_RL}
\end{figure}

\section{Exploring AdaLin's Behavior}
\vspace{-1mm}
\label{exp:neuron-behavior}
\vspace{-2mm}
\begin{figure}[!t]
    \centering
    \resizebox{0.67\linewidth}{!}{%
        \begin{minipage}{\linewidth}
            % First row: Random Label MNIST
            \begin{subfigure}[b]{0.38\linewidth}
                \centering
                \includegraphics[width=\linewidth]{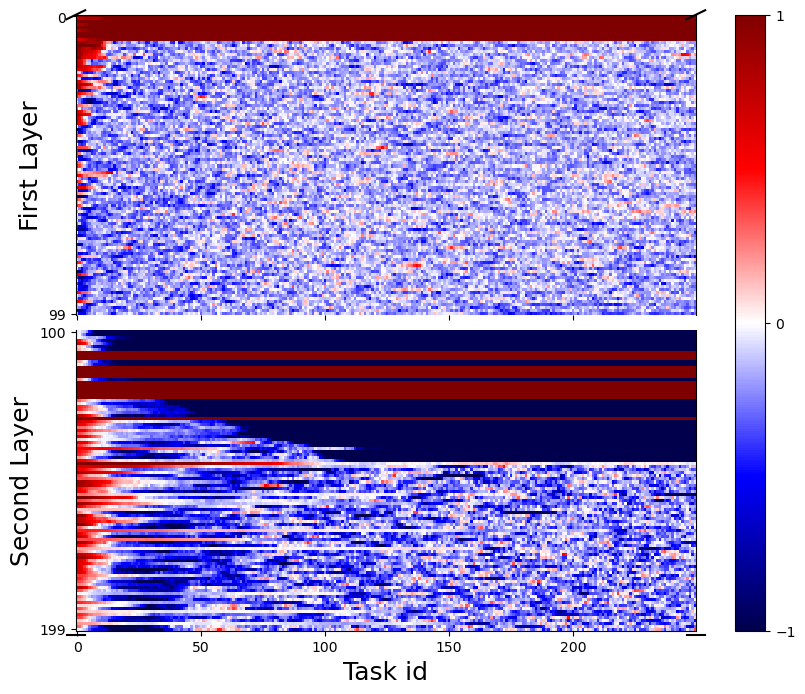}
                \caption{Learned $\alpha$ values for Random Label MNIST.}
                \label{fig:alpha_random_MNIST}
            \end{subfigure}
            \hfill
            \begin{subfigure}[b]{0.38\linewidth}
                \centering
                \includegraphics[width=\linewidth]{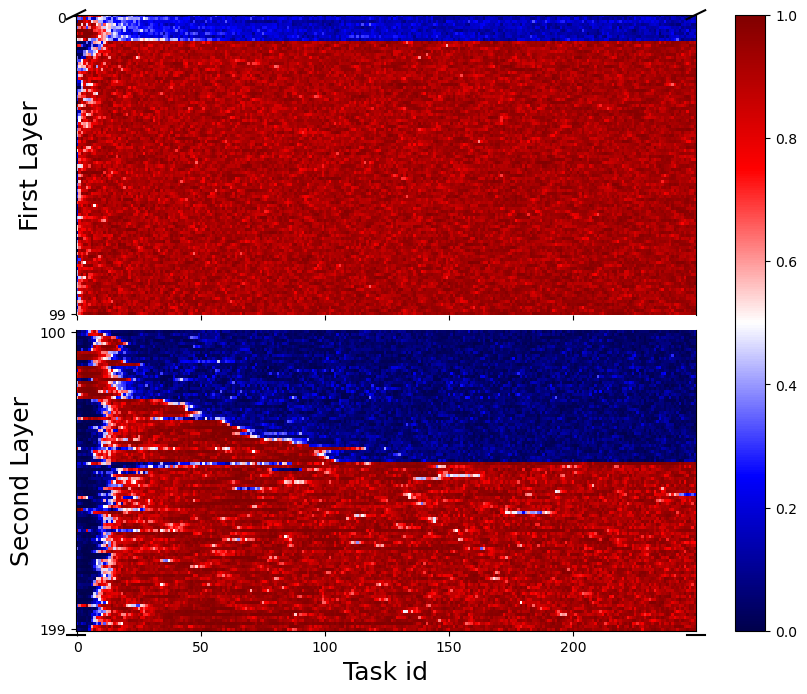}
                \caption{Average saturation $g(x)$ for Random Label MNIST.}
                \label{fig:gx_random_MNIST}
            \end{subfigure}
            
            \vspace{1em}
            
            % Second row: Random Label CIFAR-10
            \begin{subfigure}[b]{0.38\linewidth}
                \centering
                \includegraphics[width=\linewidth]{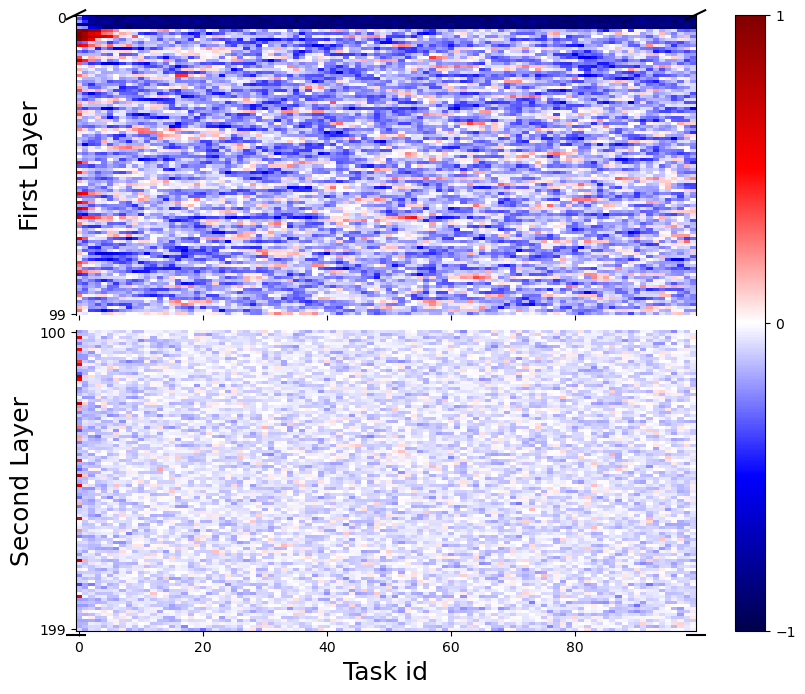}
                \caption{Learned $\alpha$ values for Random Label CIFAR-10.}
                \label{fig:alpha_random_cifar10}
            \end{subfigure}
            \hfill
            \begin{subfigure}[b]{0.38\linewidth}
                \centering
                \includegraphics[width=\linewidth]{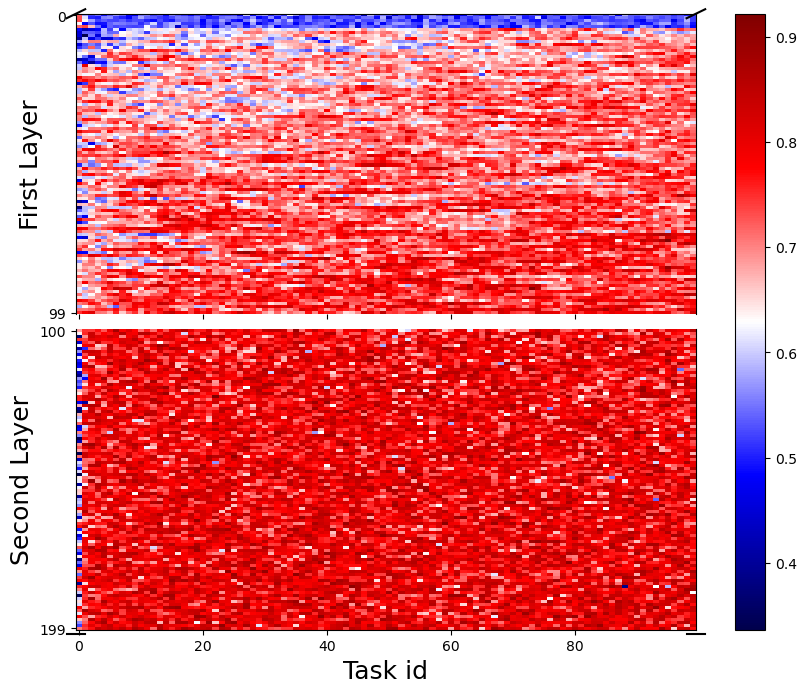}
                \caption{Average saturation $g(x)$ for Random Label CIFAR-10.}
                \label{fig:gx_random_cifar10}
            \end{subfigure}
        \end{minipage}
    }
    
    \caption{Comparison of learned $\alpha$ values (left column) and the saturation rate $g(x)$  averaged over the dataset (right column) across layers in a two-layer MLP for two continual learning tasks. Each row corresponds to one neuron and the rows are sorted from low to high saturation rates for easier interpretation.
    In the Random Label MNIST task (top row), most neurons in the first layer exhibit high saturation (high $g(x)$) and learn small negative $\alpha$ values, while a subset of unsaturated neurons converges to extreme $\alpha$ values early on and remains unchanged. In the Random Label CIFAR-10 task (bottom row), a similar pattern is observed in the second layer, with the first layer showing that more saturated neurons tend to choose more negative $\alpha$ values, whereas less saturated neurons settle at $\alpha=-1$.}
    \label{fig:alpha_saturation}
\end{figure}

To gain insight into how \texttt{AdaLin} modulates the degree of nonlinearity, we analyzed both the learned $\alpha$ values and the average saturation rate in two tasks that exhibited severe loss of plasticity for the fully continual baseline: Random Label MNIST and Random Label CIFAR-10. We measured saturation by computing the average $g(x)$ (introduced in Equation \ref{eq:DALIN}, with $\phi$ being ReLU) for each neuron at the end of each task over all training data. Figure~\ref{fig:alpha_saturation} shows the learned $\alpha$ values and average saturation per layer (in a two-layer MLP) across different tasks. For better visualization and discernibility, we sort the neurons in each of the two layers from lowest to highest average saturation rate (averaged over all tasks).

\textbf{Random Label MNIST.} In the first layer, most neurons show high $g(x)$ (red areas of the right plots), indicating saturation. These neurons learn small negative $\alpha$ values, effectively behaving like $\alpha \cdot \mathrm{ReLU}(-x) + \mathrm{ReLU}(x)$ with $\alpha$ slightly below zero. 
By contrast, a subset of neurons in the blue region of the heatmap maintains near-zero saturation and learns extreme $\alpha$ values early on.
Because these neurons remain unsaturated, their $\alpha$ parameters receive minimal gradient updates, causing them to stay fixed in subsequent tasks. We can also observe that the number of unsaturated neurons grows progressively throughout the sequence of tasks, as indicated by the widening of the blue region across tasks. 

\textbf{Random Label CIFAR-10.} Here, the second layer exhibits the highest saturation (again visible in the red areas), and most neurons converge to small negative $\alpha$. In the first layer, neurons with higher saturation similarly adopt negative $\alpha$, while those in the cooler (blue) region have lower saturation and quickly settle at $\alpha = -1$ early in training, remaining at this value thereafter.

\textbf{Common Patterns.}
Interestingly in both tasks, we observe some shared behavior. First, we can see that the majority of neurons experience high levels of saturation and utilize the linear portion of \texttt{AdaLin}. Moreover, heavily saturated neurons tend to favor small negative $\alpha$ values as indicated by the predominantly blue parts of the left plots of Fig. \ref{fig:alpha_saturation}.
On the other hand, unsaturated neurons settle into extreme $\alpha$ values and seldom change thereafter.
These observations underscore \texttt{AdaLin}’s adaptive mechanism: \textit{neurons that approach saturation inject a small linear component, whereas those that remain unsaturated rely more on their nonlinear behavior.}

\section{Ablation Study}

\subsection{Granularity Analysis}

We investigate how the granularity of the learnable modulation factor $\alpha$ affects performance, and wether it is necessary to make the parameter $\alpha$ learnable for each neuron. Specifically, we consider three cases: (1) $\alpha_i^j$ is learned separately for each neuron (Neuron Level), (2) $\alpha^j$ is shared among all neurons within a layer (Layer Level), and (3) a single $\alpha$ is shared across all neurons in the network (Network Level). As shown in Figure \ref{fig:granularity}, achieving optimal performance and fully mitigating plasticity loss is only possible when each neuron learns its own injection parameter. In particular, in Random Label MNIST and Random Label CIFAR 10, the Layer Level variant suffers from severe plasticity loss, while in Continual CIFAR 100 and Random Label CIFAR 10, the Network Level variant also experiences significant plasticity loss.

\begin{figure}[h] % [h] means 'here' (preferred position)
    \centering
    \includegraphics[width=0.9\textwidth]{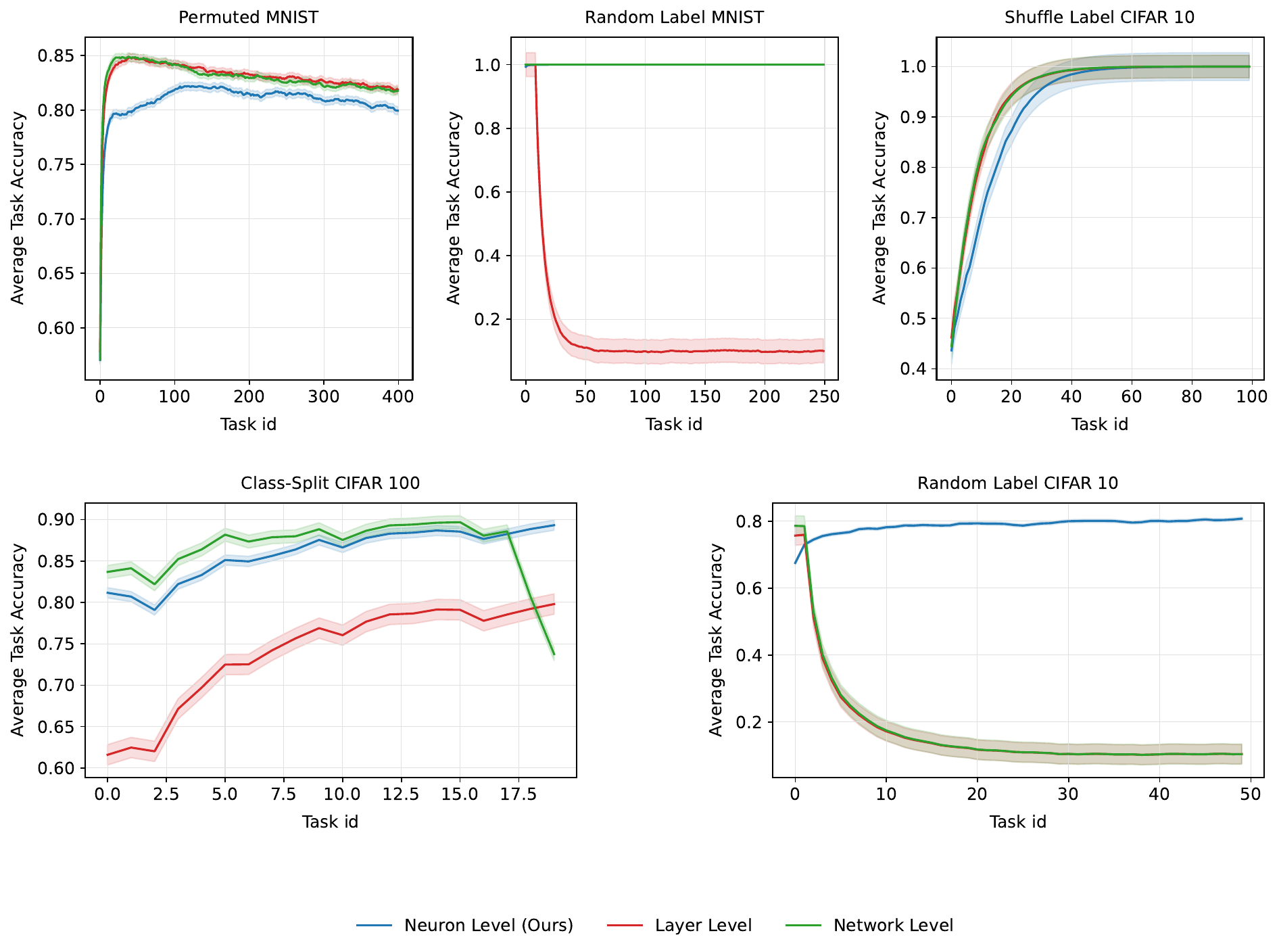}
    \caption{Impact of granularity in the learnable modulation parameter $\alpha$. The figure compares three settings: (a) Neuron-level, where each neuron learns its own $\alpha$; (b) Layer-level, where a single $\alpha$ is shared among all neurons in a layer; and (c) Network-level, where one $\alpha$ is applied across the entire network. Results on Random Label MNIST and Random Label CIFAR-10 reveal that only neuron-level adaptation successfully mitigates plasticity loss, whereas layer-level and network-level sharing lead to significant performance degradation.}

    \label{fig:granularity}
\end{figure}

\vspace{-1mm}
\subsection{why the gating mechanism matters}
\label{apendix:gx}
% \vspace{-2mm}

% In this section, we explore the question: \textit{How does using a gating mechanism $g(x)$ affect plasticity loss, and is it essential for achieving good performance?} To answer this, we compare \texttt{AdaLin} with its variant where the activation function is defined as $f_i(x) = \phi(x) + \alpha_i x$ and $\phi(x)=\text{ReLU(x)}$. As shown in Figure \ref{fig:g_matter}, the variant with $g(x)$ achieves higher training accuracy on Continual CIFAR-100 and demonstrates more stable behavior in Random Label MNIST and Shuffle Label CIFAR-10. These results suggest that it is more effective to introduce linearity only when the nonlinear component is saturated. This implies that always adding a linear component (such as a simple residual connection) does not perform as well as a gated residual connection.
%  --- 

In this section, we explore the question: \textit{How does using a gating mechanism \(g(x)\) affect plasticity loss, and is it essential for achieving good performance?} To answer this, we compare \texttt{AdaLin} with its variant where the activation function is defined as 
\[
f_i(x) = \phi(x) + \alpha_i\,x,
\]
for ReLU and GeLU as $\phi$. As shown in Figure~\ref{fig:g_matter}, When $\phi$ is ReLU, the variant with \(g(x)\) achieves higher training accuracy in all settings. Moreover, when $\phi$ is GELU, Notably, in the Random Label MNIST experiment, the variant without \(g(x)\) exhibits a pronounced loss of plasticity. These results further suggest that it is more effective to introduce linearity only when the nonlinear component is saturated, implying that always adding a linear component (such as a simple residual connection) does not perform as well as a gated residual connection.

\begin{figure}[!t] % [h] means 'here' (preferred position)
    \centering
    \includegraphics[width=0.9\textwidth]{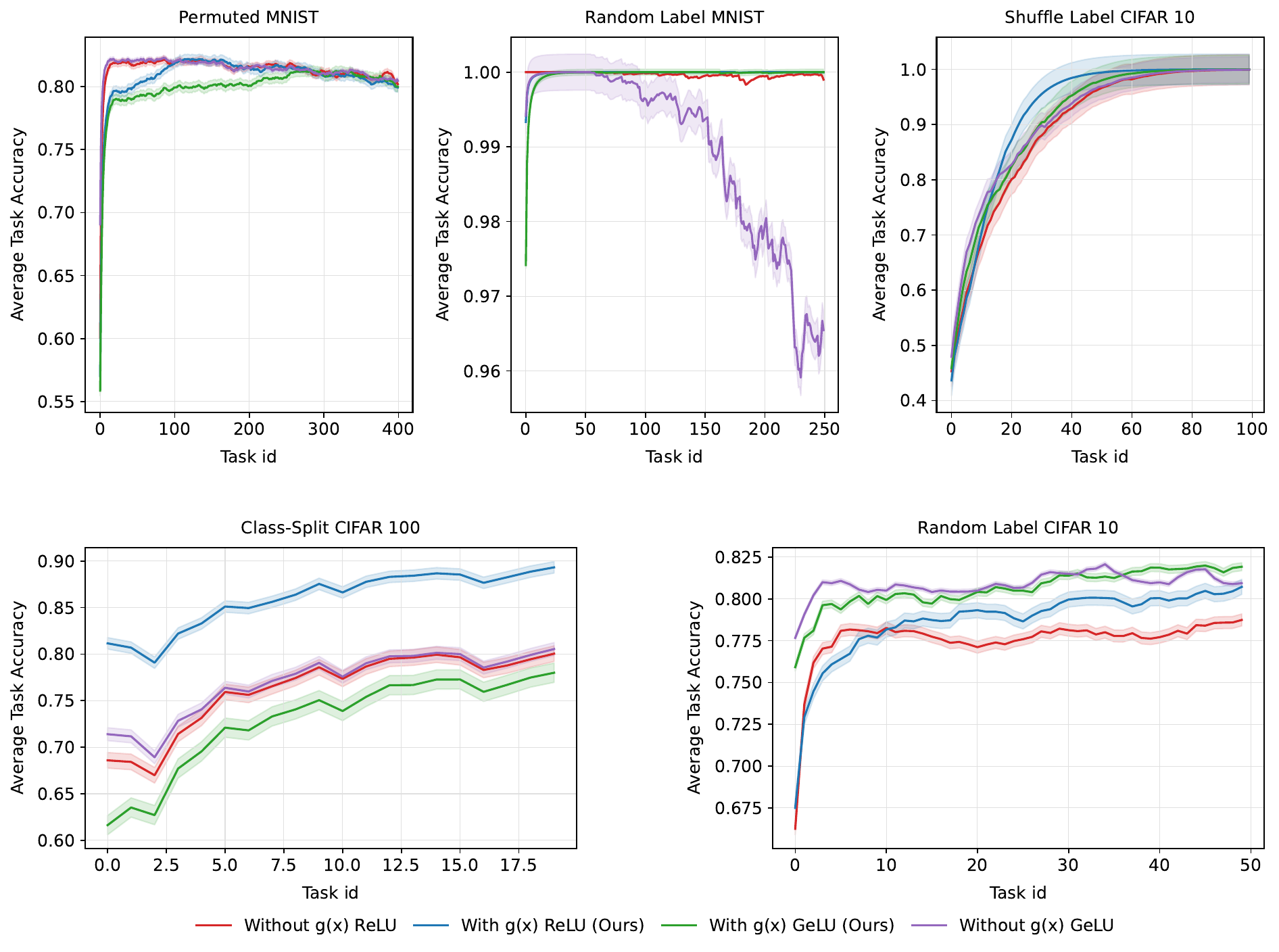}
    % \vspace{-mm}
    \caption{
    Comparison of \texttt{AdaLin} with and without a gating mechanism \(g(x)\) under both ReLU and GELU activations. The gating approach (labeled ``With \(g(x)\)'') yields higher training accuracy on all benchmarks when the activation function is ReLU. Moreover, in the Random Label MNIST experiment, the GELU variant without \(g(x)\) shows a pronounced loss of plasticity, underscoring the importance of selectively introducing linearity only when the nonlinear component is saturated.
    }
    \label{fig:g_matter}
\end{figure}

\subsection{Different choices of Gating Function}\label{appendix:lin}

To evaluate the robustness of AdaLin with respect to the choice of gating function, we tested two additional gating functions: a linear and a quadratic form.

$$g^{\text{linear}}(x) = 1 - \left( \frac{|\phi'(x)|}{L}\right)$$

$$g^{\text{quadratic}}(x) = 1 - \left( \frac{|\phi'(x)|}{L}\right)^2$$

In the above, $\phi(x)$ is the base activation function and $L$ is the Lipschitz  constant of $\phi(x)$ . As shown in Figure \ref{fig:diff_g_x}, the differences across all gating functions were not substantial. This suggests that the specific form of the gating function is less critical, as long as it satisfies key properties such as boundary conditions and monotonicity, mentioned in Section \ref{sec:adalin}.

\begin{figure}[!t] % [h] means 'here' (preferred position)
    \centering
    \includegraphics[width=0.9\textwidth]{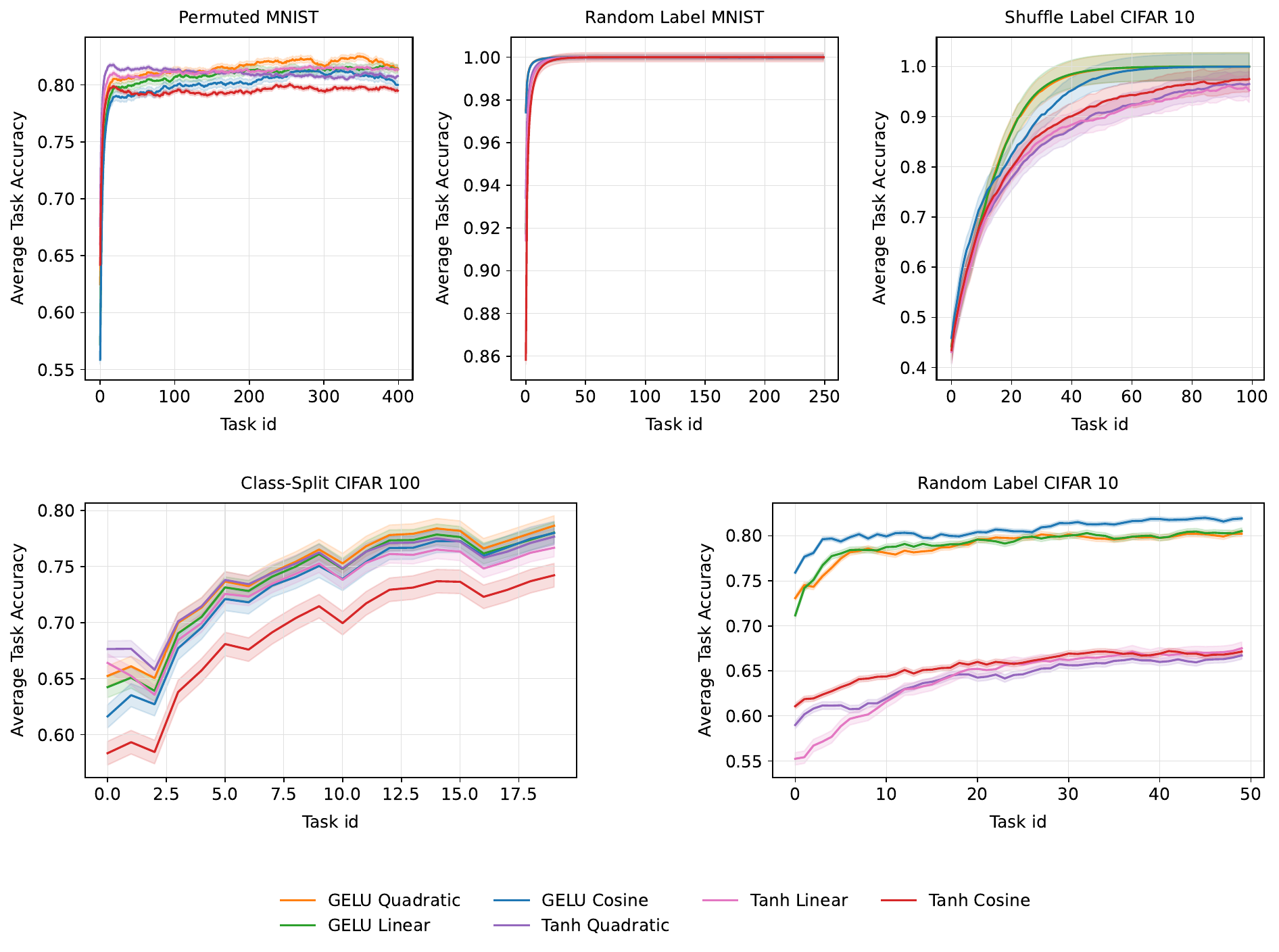}
    \caption{Performance of \texttt{AdaLin} across multiple tasks (Permuted MNIST, Random Label MNIST, Shuffle Label CIFAR-10, Class-Split CIFAR-100, and Random Label CIFAR-10) when using different gating function variants (linear, quadratic, and cosine). The differences in performance among all tested gating forms remain small, suggesting that the specific functional form of the gate is not crucial as long as it satisfies the key requirements of boundary conditions and monotonicity discussed in Section~\ref{sec:adalin}.}

    % \vspace{-mm}
    \label{fig:diff_g_x}
\end{figure}

\subsection{interpolation isn't enough}

In addition to leveraging fixed deep Fourier features in \cite{lewandowski2024plastic}, the paper introduces the concept of $\alpha$-linearization as a means to systematically blend linearity with nonlinearity within activation functions. Specifically,  $\alpha$-linearization redefines an activation function $\phi$ as
\begin{equation}
\phi_{\alpha}(x) = \alpha x + (1 - \alpha) \phi(x),
\label{eq:interpolate}
\end{equation}
where the parameter $\alpha$ governs the trade-off between purely linear behavior ($\alpha$ = 1) and the original nonlinear response ($\alpha$ = 0). A critical challenge is the sensitivity of performance to the precise tuning of $\alpha$. Even with optimal tuning, a high $\alpha$ can result in overly linear representations that slow down training, while a low $\alpha$ may fail to adequately preserve plasticity.  This naturally raises the question of what happens if $\alpha$ is made learnable for each neuron, allowing each unit to dynamically interpolate between linear and nonlinear behavior. Such a tunable interpolation could help address issues like saturation and unit linearization that often hinder continual learning. In order to validate this, we use Equation \ref{eq:interpolate} as the activation function of each neuron with $\phi(.)$ being and the $\alpha$ parameter being learnable per neuron. In order to bound $\alpha$ between 0 and 1, we pass the learned $\alpha$ through a sigmoid. The results of this approach on the plasticity benchmarks are reported in Figure \ref{fig:interpolation}
. As it can be seen, This approach can not avoid loss of plasticity, and the model completely fails to preserve it's average task accuracy in the Random Label Cifar 10 and Random Label MNIST tasks. This suggests that while interpolation inherently adds linearity to the network, it should be noted that (1) the optimal magnitude of added linearization to the neuron might not necessarily be between 0 and 1, and (2) the gating mechanism that indicates the conditions on when the linearity should be added to the neuron plays an important role. 
\begin{figure}[htb] % [h] means 'here' (preferred position)
    \centering
    \includegraphics[width=0.9\textwidth]{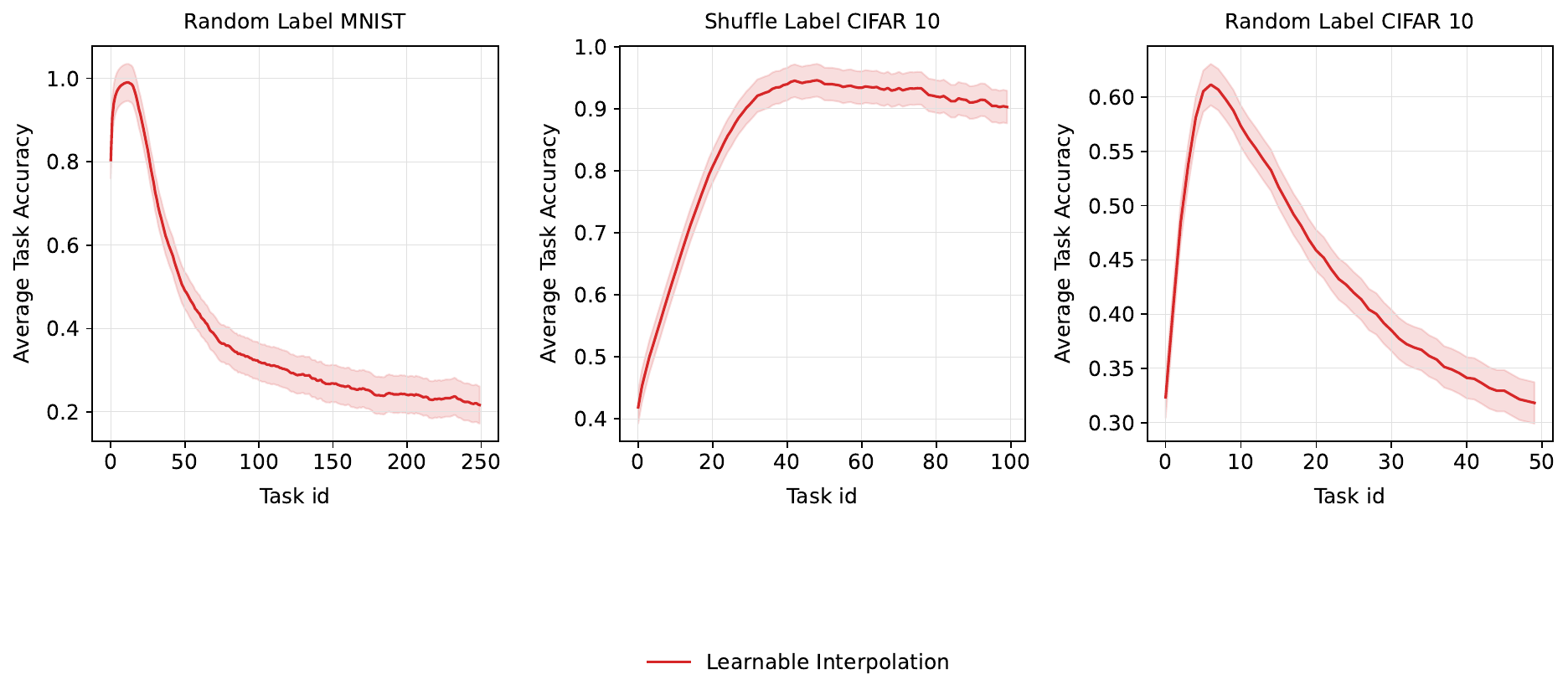}
        \caption{Evaluation of learnable interpolation using ReLU across plasticity benchmarks. The method introduces a neuron-specific, learnable parameter $\alpha$ to balance linear and nonlinear behavior dynamically. However, as shown in the results, this approach fails to prevent plasticity loss, particularly in Random Label MNIST and Random Label CIFAR-10, where average task accuracy declines significantly over time. This suggests that while interpolation introduces controlled linearity, effective plasticity preservation may require alternative strategies such as optimizing the range of $\alpha$ beyond $[0,1]$ or incorporating a gating mechanism to regulate when and how linearization is applied.}
    \label{fig:interpolation}
\end{figure}

\section{Analyzing Different Criteria}

\subsection{Parameter Overhead Analysis}
\label{appendix:overhead}
In our method, we add one learnable parameter per neuron in the MLP and one per convolutional filter in the CNN. Below we detail the calculations for each network. Table \ref{tab:overhead} summarizes the results.
\paragraph{MLP.}
For a two-layer MLP with 100 hidden units and a 10-unit output (assuming input dimension $d=784$):
\[
\begin{array}{rcl}
\text{Layer 1:} & 100(d+1) &= 100\times785 = 78\,500, \\
\text{Layer 2:} & 100(100+1) &= 10\,100, \\
\text{Output:} & 10(100+1) &= 1\,010.
\end{array}
\]
Total parameters: $78\,500+10\,100+1\,010 = 89\,610$. Added parameters (one per neuron): $100+100+10 = 210$. Overhead: 
\[
\frac{210}{89\,610}\times100 \approx 0.23\%.
\]

\paragraph{CNN.}
For the CNN:
\begin{itemize}
    \item Conv1: $(5\times5\times3)\times16+16 = 1\,216$,
    \item Conv2: $(5\times5\times16)\times16+16 = 6\,416$,
    \item FC1: $400\times64+64 = 25\,664$,
    \item FC2: $64\times100+100 = 6\,500$.
\end{itemize}
Total parameters: $1\,216+6\,416+25\,664+6\,500 = 39\,796$. Added parameters (one per filter in each conv layer): $16+16 = 32$. Overhead:
\[
\frac{32}{39\,796}\times100 \approx 0.08\%.
\]

\begin{table}[H]
\centering
\begin{tabular}{lccc}
\toprule
Network & Added Parameters & Total Parameters & Overhead (\%) \\
\midrule
MLP & 210 & 89\,610 & 0.23\% \\
CNN & 32 & 39\,796 & 0.08\% \\
\bottomrule
\end{tabular}
\caption{Parameter overhead due to the additional learnable parameter per neuron/filter.}
\label{tab:overhead}
\end{table}

\subsection{Analyzing Entropy}

Sign entropy is a measure of activation diversity in neural networks, capturing how frequently a neuron’s output switches between positive and negative values across different inputs. A neuron with high sign entropy exhibits a balanced mix of positive and negative activations, suggesting that it remains trainable and responsive to new tasks. In contrast, low sign entropy indicates that a neuron consistently outputs values of the same sign (always positive or negative), which can be a sign of linearization or saturation of that neuron. To evaluate the role of sign entropy in our model’s ability to learn continually, we examined the entropy of neuron activations at the end of each task in the experiments in Section  Section \ref{exp:plastic-benchmarks}. Specifically, at the end of the task, we computed sign entropy by passing the training data through the model and estimating the entropy of each neuron’s activations. This allowed us to track how the behavior of activation signs evolved over multiple tasks. We then plotted the average entropy over all neurons in the network to examine if sign entropy might be correlated with loss of plasticity. We repeated the same procedure for the \texttt{Scratch} (where we expect no increase in entropy) and \texttt{Baseline} to measure how effective Adalin is. The results of are depicted in Figure \ref{fig:stats_entropy}.

Across all benchmark settings, the \texttt{Baseline} (orange) consistently exhibits a steep decrease in sign entropy, even in tasks where the loss of plasticity wasn't observed like Continual Cifar 100. In contrast, \texttt{AdaLin} maintains sign entropy levels closer to those of the \texttt{Scratch} model. Notably, in some cases, such as Permuted MNIST and Continual CIFAR-100, \texttt{AdaLin} even has a sign entropy greater than the \texttt{Scratch} model, while in tasks like Random Label MNIST and Shuffle Label CIFAR10 the sign entropy of \texttt{AdaLin} is lower than that of the \texttt{Scratch}, despite not observing loss of plasticity. These discrepancies might indicate that sign entropy alone might not be an adequate measure for loss of plastity.

\begin{figure}[h] % [h] means 'here' (preferred position)
    \centering
    \includegraphics[width=0.9\textwidth]{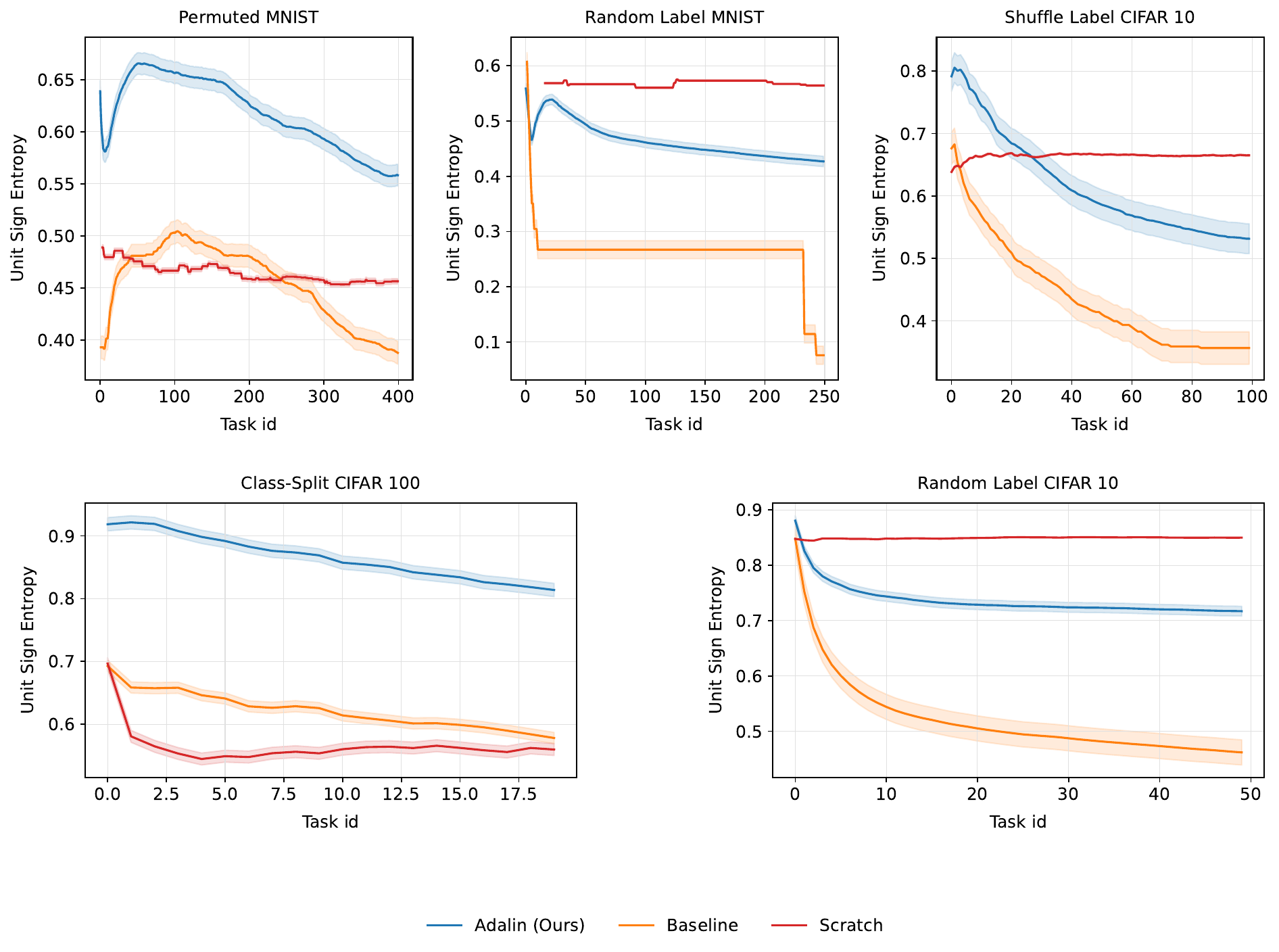}
    \caption{Evolution of sign entropy across tasks in different plasticity benchmarks. The \texttt{Baseline} (orange) exhibits a steep decline in sign entropy, indicating a significant loss of activation diversity and plasticity over time. In contrast, Adalin (blue)  maintains activation diversity at levels closer to or even exceeding the \texttt{Scratch} model (red).}
    \label{fig:stats_entropy}
\end{figure}

\subsection{A Study of S-Rank and Weight Norms}

In this section, we analyze two other factors that, based on recent work, might be relevant to plasticity loss in neural networks:
\begin{itemize}
    \item \textbf{Weight Norm:} The norm of a network’s weight parameters measures the overall scale of the model’s learned weights. Large weight magnitudes can inhibit adaptation, particularly when using adaptive optimizers like Adam, where step sizes are dynamically adjusted. 

    \item \textbf{Feature Rank (SRank):} The effective rank of the network’s weight matrices can serve as a proxy for the diversity of learned features. A model with high feature rank retains a rich representational space, allowing it to adapt flexibly to new tasks. However, if feature rank collapses—where the weight matrices become nearly low-rank—learning capacity is significantly reduced. This collapse may occur due to over-regularization or continual training on non-stationary data, leading to representations that are no longer expressive enough to accommodate new information.
\end{itemize}

Figures \ref{fig:stats_norm} and \ref{fig:stats_srank} illustrate the weight norm and feature rank of the model across all plasticity benchmarks for the analyzed experiments in Section \ref{exp:plastic-benchmarks}. In Figure \ref{fig:stats_norm}, the weight norm trends reveal that \texttt{AdaLin} exhibits a gradual and controlled increase. Additionally, the \texttt{Deep Linear} model, which does not suffer from plasticity loss across tasks, also demonstrates a steady rise in weight norm. Surprisingly, the \texttt{Baseline} model shows an increase in weight norm across all tasks—both those where it inherently suffers from plasticity loss (i.e., Random Label MNIST, Random Label CIFAR-10, and Random Shuffle CIFAR-10) and those where it does not (i.e., Permuted MNIST and Continual CIFAR-100). This observation suggests that weight norm alone may not strongly correlate with plasticity loss.

In Figure \ref{fig:stats_srank}, it can be observed that across all experiments, \texttt{AdaLin} maintains a feature rank comparable to that of models trained from \texttt{Scratch}, which may explain its ability to preserve plasticity. However, the \texttt{Baseline} also does not exhibit a steep decline in any setting, making it unclear whether SRank can be reliably correlated with plasticity loss. Surprisingly, ]\texttt{CReLU} maintains the highest SRank across all settings, even in those where it experiences significant plasticity loss (e.g., Continual CIFAR-100).
\begin{figure}[h] % [h] means 'here' (preferred position)
    \centering
    \includegraphics[width=0.9\textwidth]{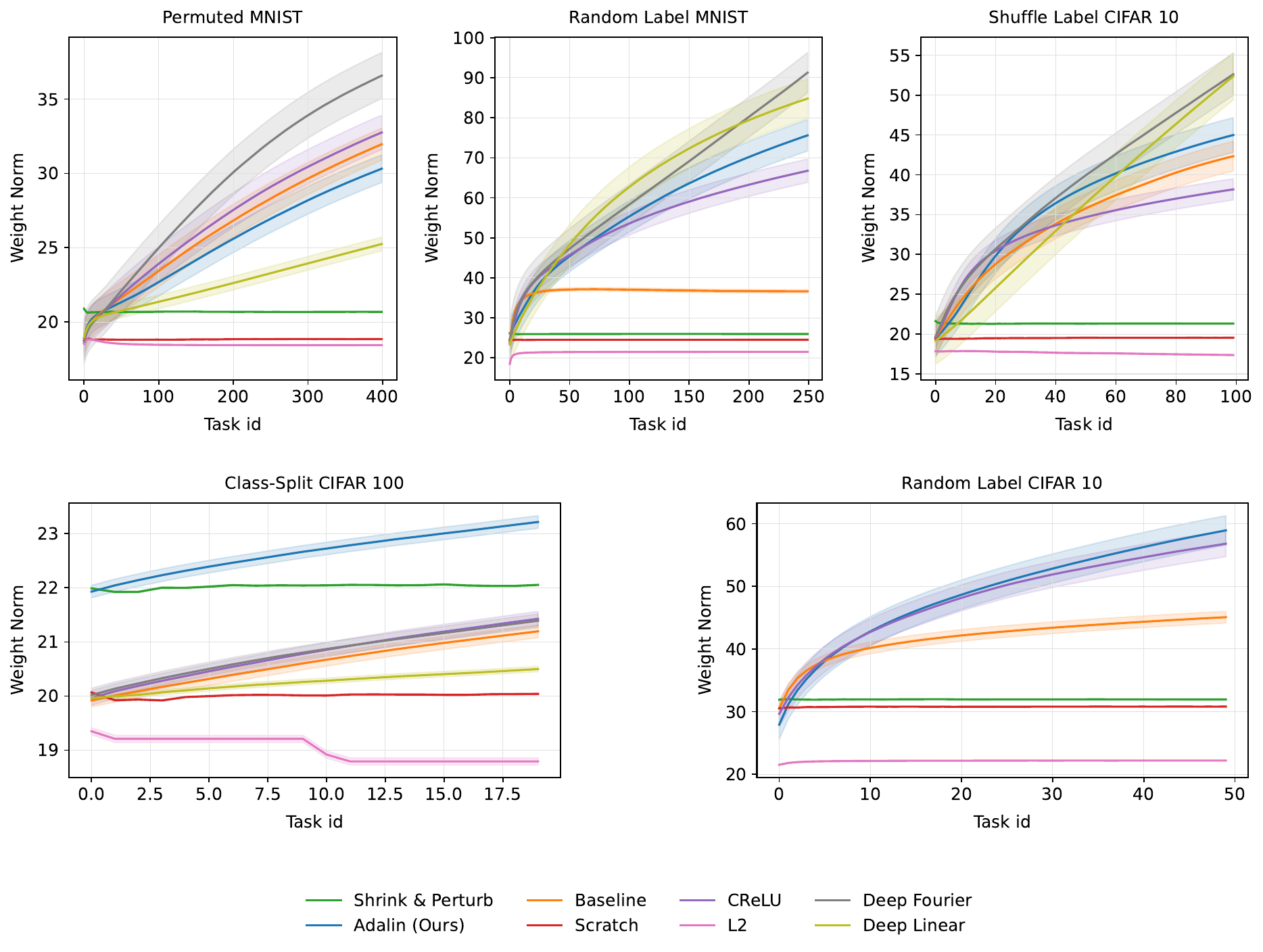}
    \caption{Evolution of weight norm across tasks for different plasticity benchmarks. \texttt{AdaLin} (blue) exhibits a gradual and controlled increase in weight norm, similar to the \texttt{Deep Linear} model (yellow), which is known to maintain plasticity across tasks. Surprisingly, the \texttt{Baseline} model (orange) shows an increase in weight norm across all tasks, regardless of whether it suffers from plasticity loss. This suggests that weight norm alone may not serve as a strong indicator of plasticity degradation.}
    \label{fig:stats_norm}
\end{figure}

\begin{figure}[h] % [h] means 'here' (preferred position)
    \centering
    \includegraphics[width=0.9\textwidth]{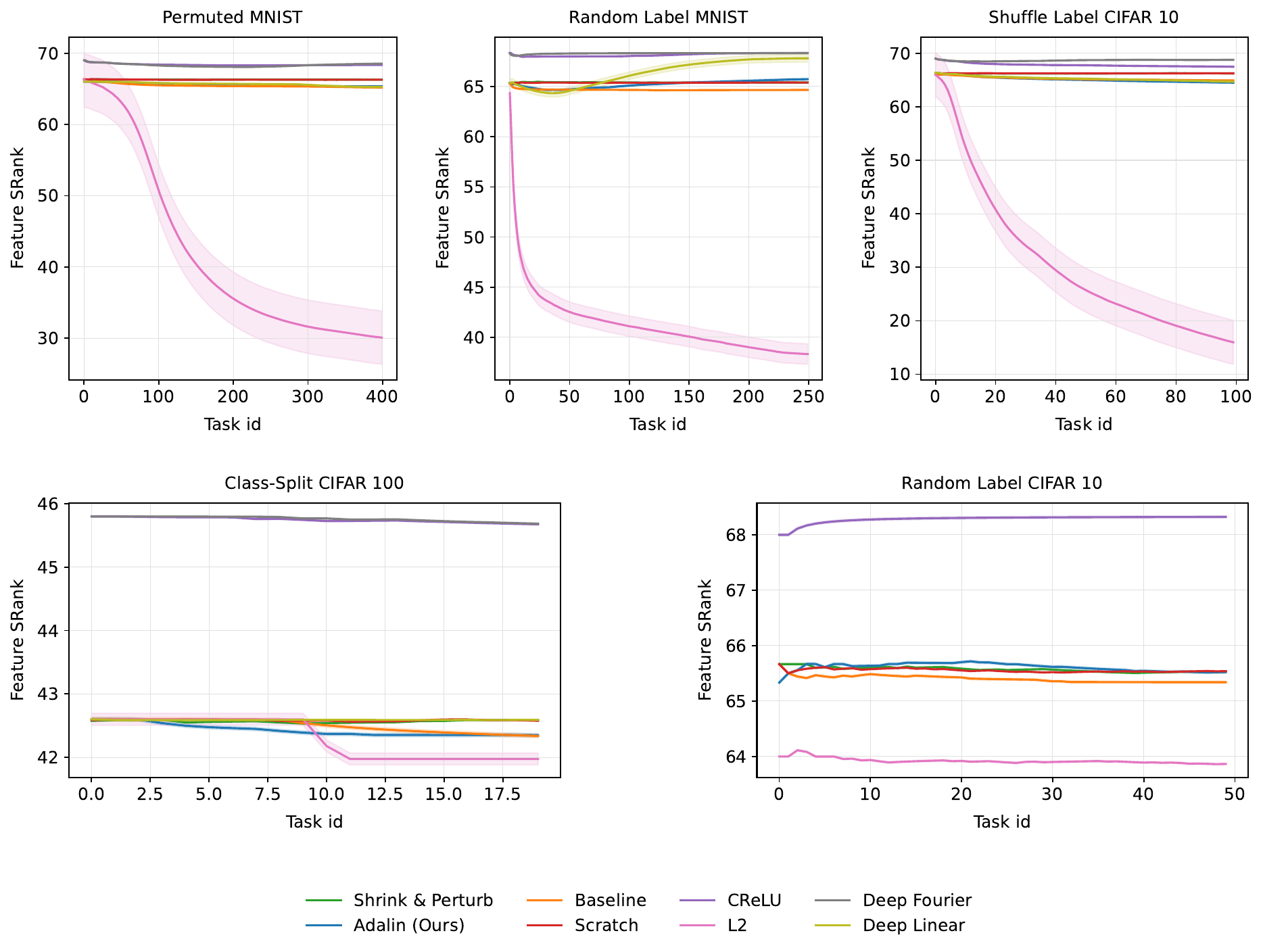}
    \caption{ Evolution of feature rank (SRank) across tasks for different plasticity benchmarks. \texttt{AdaLin} (blue) maintains a feature rank comparable to that of models trained from \texttt{Scratch} (red), which may contribute to its ability to preserve plasticity. Interestingly, the \texttt{Baseline} (orange) does not exhibit a steep decline in any setting, raising questions about the correlation between SRank and plasticity loss. Notably, \texttt{CReLU} (purple) consistently achieves the highest SRank across all settings, even in scenarios where it suffers from severe plasticity loss (e.g., Continual CIFAR-100), suggesting that a high feature rank alone does not necessarily imply preserved plasticity.}
    \label{fig:stats_srank}
\end{figure}

\section{Neuro-Inspired Design of \texttt{AdaLin}}

\label{sec:neuro_inspired}

\textbf{Adaptive Gain Control in Biological Neurons:}
Adaptive gain control allows neurons to modulate their sensitivity or response gain in response to varying stimuli and brain states \cite{Ferguson2020, Cheadle2014, McGinley2015}. For example, research by Ferguson and Cardin \cite{Ferguson2020} highlights how cortical neurons adjust their gain through cellular mechanisms such as GABAergic inhibition and membrane potential fluctuations. Cheadle et al. \cite{Cheadle2014} showed that human perceptual decision-making involves adaptive tuning of neural response functions, allowing more weight to consistent information. McGinley et al. \cite{McGinley2015} demonstrated that neuronal gain in sensory cortex follows an inverted-U relationship with arousal, where moderate arousal optimizes responsiveness.

\textbf{Dendritic Computation and Local Non-Linearities:}
Neurons integrate inputs through their dendrites, which perform complex local computations and can act as non-linear subunits \cite{Gambino2014, Bittner2015, Gidon2020}. Gambino et al. \cite{Gambino2014} showed that dendritic plateau potentials in cortical neurons drive synaptic plasticity independently of somatic spikes. Bittner et al. \cite{Bittner2015} found that hippocampal neurons generate dendritic spikes only with coincident input, effectively functioning as AND-gates for memory encoding. Gidon et al. \cite{Gidon2020} reported that human cortical neurons exhibit graded dendritic action potentials, enabling single neurons to classify non-linearly separable input patterns.

\textbf{Adaptive Non-Linearity in Neural Circuits:}
Neural circuits can adapt their non-linear response properties through circuit-level mechanisms \cite{Lovett2014, Palacios2021, Rao2017}. Lovett-Barron et al. \cite{Lovett2014} identified a gating mechanism in the hippocampus that controls dendritic non-linearity based on behavioral context. Palacios et al. \cite{Palacios2021} demonstrated experience-dependent increases in dendritic plateau events in sensory cortex. Rao and Geffen \cite{Rao2017} showed how different types of inhibitory interneurons modulate the non-linear responses of cortical neurons by scaling gain or shifting response thresholds.

These neuroscientific insights inspired our \texttt{AdaLin} method's dynamic adaptation of activation functions. By incorporating a tunable linear component into any activation function when it saturates, \texttt{AdaLin}  enhances the suitability of artificial neurons for continual learning, mirroring the adaptive behaviors observed in biological systems.

\section{Miscellaneous}
\subsection{Recovery of PReLU from \texttt{AdaLin} with ReLU}
\label{sec:appendix_prelu}

Here, we show that if the base non-linearity \(\phi(x)\) is instantiated as ReLU, \texttt{AdaLin} reduces to the Parametric ReLU (PReLU). Recall that \texttt{AdaLin} is defined as
$$
f_i^j(x) = \phi(x) + \alpha_i^j\, x\, [ \cos\left(\frac{\pi}{2}\cdot \frac{|\phi'(x)|}{L}\right)]_{\text{sg}},
$$
where \(\alpha_i^j\) is a learnable scaling factor, \(L\) is the Lipschitz constant of \(\phi\), and \(\square\) denotes stop-gradient.

For ReLU, we have:
\[
\phi(x) = \begin{cases}
0, & x < 0,\\[1mm]
x, & x \ge 0,
\end{cases}
\quad \text{and} \quad
\phi'(x) = \begin{cases}
0, & x < 0,\\[1mm]
1, & x > 0.
\end{cases}
\]
Since ReLU is 1-Lipschitz, we set \(L = 1\).

\textbf{For \(x < 0\):}  
Here, \(\phi(x) = 0\) and \(|\phi'(x)| = 0\), so the gating function becomes
\[
g(x) = \cos\left(\frac{\pi}{2}\cdot \frac{0}{1}\right) = \cos(0) = 1.
\]
Thus,
\[
f_i^j(x) = 0 + \alpha_i^j\, x = \alpha_i^j\, x,
\]
which corresponds to the negative part of the PReLU activation.

\textbf{For \(x \ge 0\):}  
In this regime, \(\phi(x) = x\) and \(|\phi'(x)| = 1\), hence
\[
g(x) = \cos\left(\frac{\pi}{2}\cdot \frac{1}{1}\right) = \cos\left(\frac{\pi}{2}\right) = 0.
\]
Therefore,
\[
f_i^j(x) = x + \alpha_i^j\, x \cdot 0 = x,
\]
which is exactly the ReLU function for non-negative inputs.

In summary, with \(\phi(x) = \mathrm{ReLU}(x)\), \texttt{AdaLin} yields
\[
f_i^j(x) = \begin{cases}
\alpha_i^j\, x, & x < 0,\\[1mm]
x, & x \ge 0,
\end{cases}
\]
thereby recovering the PReLU formulation.

\subsection{Extension to Convolutional Neural Networks}
\label{sec:appendix_cnn}

In this appendix, we briefly describe the extension of \texttt{AdaLin} to Convolutional Neural Networks (CNNs). For a convolutional layer, let \(h_c^j(u,v)\) denote the post-activation output at spatial location \((u,v)\) for channel \(c\) in the \(j^{\text{th}}\) layer. The convolution operation is given by
\[
h_c^j(u,v) = f_c^j\Big((\mathbf{W}_c^j * \mathbf{h}^{j-1})(u,v)\Big),
\]
where \(\mathbf{W}_c^j\) is the convolution kernel and \(\mathbf{h}^{j-1}\) the preceding feature map.

Adapting \texttt{AdaLin} to CNNs, the activation function is defined as
$$
f_c^j(x) = \phi(x) + \alpha_c^j \, x\, \left[ \cos\left(\frac{\pi}{2} \cdot \frac{|\phi'(x)|}{L}\right)\right]_{\text{sg}},
$$

where \(\alpha_c^j\) is a learnable scaling factor (shared across all spatial locations within channel \(c\) in layer \(j\)), \(\phi(\cdot)\) is the base non-linear function, \(L\) is its Lipschitz constant, and \(\square\) denotes the stop-gradient operation. Notably, when \(\phi(\cdot)\) is set to ReLU, this formulation recovers the PReLU variant.

This concise formulation inherits the adaptive residual mechanism from the main text, ensuring improved gradient flow and preserved plasticity in CNNs.

\subsection{Determining the Lipschitz Constant for Common Non-linearities}
\label{sec:appendix_lipschitz}

Recall that our gating function is defined as
\[
g(x) = \cos\left(\frac{\pi}{2} \cdot \frac{|\phi'(x)|}{L}\right),
\]
where \(L\) is the Lipschitz constant of the activation function \(\phi(x)\). Below we describe the choice of \(L\) for several popular activations:

\begin{itemize}
    \item \textbf{ReLU:} The ReLU activation is defined as 
    \[
    \mathrm{ReLU}(x) = \max(0,x).
    \]
    Its derivative is either \(0\) or \(1\), so the maximum \(|\phi'(x)|\) is \(1\), which implies \(L = 1\).

    \item \textbf{Tanh:} The hyperbolic tangent function is given by
    \[
    \tanh(x) = \frac{e^x - e^{-x}}{e^x + e^{-x}},
    \]
    with derivative
    \[
    \tanh'(x) = 1 - \tanh^2(x).
    \]
    Since the maximum value of \(1 - \tanh^2(x)\) is \(1\) (attained at \(x=0\)), we set \(L = 1\).

    \item The GELU activation function is defined as:
\[
\text{GELU}(x) = x \Phi(x),
\]
where \( \Phi(x) \) is the standard Gaussian cumulative distribution function (CDF):

\[
\Phi(x) = \frac{1}{2} \left(1 + \operatorname{erf} \left(\frac{x}{\sqrt{2}}\right) \right).
\]

To find its Lipschitz constant \( L \), we compute its derivative:

\[
\frac{d}{dx} \text{GELU}(x) = \Phi(x) + x \phi(x),
\]

where \( \phi(x) \) is the standard Gaussian probability density function (PDF):

\[
\phi(x) = \frac{1}{\sqrt{2\pi}} e^{-x^2/2}.
\]

The Lipschitz constant \( L \) is given by:

\[
L = \sup_{x} \left| \Phi(x) + x \phi(x) \right|.
\]

Numerically, this supremum is approximately:

\[
L \approx 1.12.
\]

\end{itemize}

These choices ensure that the gating function is appropriately scaled to the maximum sensitivity of each activation function.

\end{document}